\documentclass[letterpaper,12pt,onecolumn]{article}  %

\newcommand{\be}{\begin{equation}}
\newcommand{\ee}{\end{equation}}
\newcommand{\beq}{\begin{equation}}
\newcommand{\eeq}{\end{equation}}
\newcommand{\bed}{\begin{displaymath}}
\newcommand{\eed}{\end{displaymath}}
\newcommand{\beqa}{\begin{eqnarray}}
\newcommand{\eeqa}{\end{eqnarray}}
\newcommand{\beqann}{\begin{eqnarray*}}
\newcommand{\eeqann}{\end{eqnarray*}}
\newcommand{\bseq}{\begin{subequation}}
\newcommand{\eseq}{\end{subequation}}

\newcommand{\ba}{\begin{array}}
\newcommand{\ea}{\end{array}}

\newcommand{\x}{{\bf x}}

\usepackage[dvips]{graphicx}
\usepackage{psfrag}
\usepackage{epsfig}
\usepackage{array}
\usepackage{amssymb}
\usepackage{subeqn}
\usepackage{setspace}
\usepackage{subfigure}
\usepackage{longtable}
\usepackage{color}
\usepackage{multirow}
\definecolor{myblue}{rgb}{0,0.627,0.776}
\definecolor{mygreen}{rgb}{0.2,0.627,0.173}
\definecolor{myred}{rgb}{0.537,0.047,0.031}

\makeatletter
\begin{document}
\date{}
\vspace{-1 cm}
\title{\large{\bf Path Placement Optimization of Manipulators Based on Energy Consumption: \goodbreak  Application to the Orthoglide 3-axis}
\author{\normalsize{{\bf Raza UR-REHMAN, \, St\'{e}phane CARO}}\\
\normalsize{{\bf Damien CHABLAT, \, Philippe WENGER}} \\
\normalsize{Institut de Recherche en Communications et
Cybern\'etique de Nantes}\\
\normalsize{UMR CNRS n$^\circ$ 6597, 1 rue de la No\"e, 44321 Nantes, France}\\
\normalsize{\tt \{ur-rehman, caro, chablat, wenger\}@irccyn.ec-nantes.fr}
}  }
\maketitle

\thispagestyle{empty}
\subsection*{Abstract}
\begin{sloppypar}
This paper deals with the optimal path placement for a manipulator based on energy consumption. It proposes a methodology to determine the optimal location of a given test path within the workspace of a manipulator with minimal electric energy used by the actuators while taking into account the geometric, kinematic and dynamic constraints. The proposed methodology is applied to the Orthoglide~3-axis, a three-degree-of-freedom translational parallel kinematic machine (PKM), as an illustrative example.
\end{sloppypar}
\begin{sloppypar}
\noindent{\bf Keywords:} Path placement, Energy consumption, Optimization, Parallel manipulators.
\end{sloppypar}
\begin{center}
\large{\bf Placement optimal de trajectoires de manipulateurs pour la minimisation de la consommation d'\'energie~: \goodbreak application \`a l'Orthoglide~3-axes}
\end{center}
\subsection*{R\'esum\'e }
\begin{sloppypar}
L'objet de cet article est le placement optimal de trajectoires en utilisant la consommation \'energ\'etique comme crit\`ere. Ce travail propose une méthodologie pour d\'eterminer la localisation optimale de la trajectoire dans l'espace de travail d'un manipulateur en minimisant la consommation d'\'energie \'electrique utilis\'ee par les moteurs, tout en tenant compte des contraintes g\'eom\'etriques, cin\'ematiques et dynamiques. La m\'ethodologie propos\'ee est appliqu\'ee \`a l'Orthoglide 3-axes, manipulateur d'architecture parall\`ele \`a trois degr\'es de libert\'e de translation.
\end{sloppypar} 
\begin{sloppypar}
\noindent{\bf Mots cl\'es:} Placement de trajectoires, Consommation \'energ\'etique, Optimisation, Manipulateur parall\`eles.
\end{sloppypar}

\section{INTRODUCTION}
Optimal trajectory planning of manipulators has been a relevant area for roboticists for many years. Indeed, several authors have worked on trajectory planning based on different optimization objectives. A review of trajectory planning techniques is given in  \cite{AtaAA2008}.
Trajectory planning deals with the determination of the path and velocity/acceleration profiles (or the time history of the robot's joints), the start and end points of the trajectory being predefined and fixed in the workspace. As a matter of fact, this approach is suitable for most of robotic applications.
A path is a continuous curve in the configuration space connecting the initial configuration of the manipulator to its final configuration \cite{RajanVT1985}. Trajectory planning usually aims at minimizing the travel distance \cite{Pires2001,Tian2003}, travel time \cite{ChanKK1993,CaoB1994,PledelP1995} and/or the energy consumed \cite{ShugenM1995,FieldG1995,Hirakawa1997, Khoukhi2007}, while satisfying several geometric, kinematic and dynamic constraints. 

Another less explored aspect of trajectory planning is the placement of a given path within the workspace. It aims at determining the optimum location of a predefined path to be followed by the end-effector of the manipulator within its workspace with respect to one or many given objective(s) and constraint(s). This path can be the shape of a component to be machined, a welded profile or an artistic/decorative profile etc. In such situations, the trajectory planner cannot alter the shape of the path but he/she can only play with the location of that path within the workspace of the manipulator in order to optimize one or several criterion(a). Such an approach can be very interesting in many robotic applications. For example, in machining, a workpiece can be better located within the workspace of the robot to perform a given operation more efficiently with respect to the energy consumed.

The path placement problem has not been extensively studied in the past. Nevertheless, some researchers proposed to solve it with respect to various optimization objectives. Several performance criteria for path location problems can be considered simultaneously (multiobjective) or individually, such as travel time, different kinetostatic performance indices (manipulability or conditioning number), kinematic performance (velocity, acceleration), avoiding obstacles, reduced wear or vibration, energy consumption etc. In the following paragraphs, a brief survey of the work of different researchers to solve the problem of path placement optimization for various applications is presented.

Nelson and Donath \cite{Nelson1987} proposed an algorithm for the optimum location of an assembly task in the manipulator workspace while taking the manipulability measure as the optimization criterion. They considered that the location of the assembly task within the workspace that results in the highest manipulability is a locally optimal position for performing the assembly. However, Aspragathos \cite{Aspragathos1996,Aspragathos2002} considered that the manipulability index and the dexterity, usually quantified by the condition number of the Jacobian matrix of the manipulator, can characterize the motion ability of the manipulator, but these criteria cannot depict the ability of a manipulator to move in a given direction. Hence they introduced a criterion to characterize the best velocity performance of the robot end-effector with the path location. They used the concept of the orientation of the manipulability ellipsoid relative to the desired path and used genetic algorithm to come up with an optimal solution.

Fardanesh et al \cite{Fardanesh1988} proposed an approach for optimal positioning of a prescribed task in the workspace of a 2\textsl{R}-manipulator. Optimal location of the task is considered to be the location that yields the minimum cycle time for the task. In another study, Feddema \cite{Fardanesh1996} formulated and solved a problem of robot base placement for minimum time joint coordinated motion within a work cell. The proposed algorithm considers only the kinematics and the maximum acceleration of each joint in order to obtain a $25\%$ cycle time improvement for a typical example. 

Hemmerle \cite{Hemmerle1991} presented an algorithm for optimum path placement of a redundant manipulator by defining a cost function related to robot joints motion and limits. The proposed approach did not consider the path as a whole but points along the entire path, hence cost function considers the performance only at the node points and not the path in-between the nodes.

Chou and Sadler \cite{Chou1993} developed an optimization technique for the optimum placement of a robotic manipulator based on the actuators torque requirements. Pamanes and Zeghloul \cite{Pamanes1989,Pamanes1991} considered multiple kinematic indices to find the optimal placement of a manipulator by specifying the path with a number of points and then assigning an optimization criterion to each point. The objective was to find the path location in order to have optimal values of all the criteria assigned to the path's points. In \cite{Pamanes1991B}, the problem of optimal placement with joint-limits and obstacle avoidance is addressed. Lately, a general formulation was presented to determine the optimal location of a path for a redundant manipulator while dealing with mono- and multi-objective problems \cite{Pamanes2008}. The goal of this research work was to keep the joint variables within their limits and to minimize the magnitude of their displacements \cite{Pamanes1991C,Pamanes1995,Zeghloul1990,Zeghloul1991,Zeghloul1993,Zeghloul1993B,Zeghloul1994}.

With a general literature survey, it comes out that although several performance indices are introduced or considered, there is very little emphasis on the dynamic aspects reflecting the energy consumption by the robot actuators. The increasing number of the robotic applications emphasize the importance of the energy saving not only to enhance efficiency but also to face the world energy problems. Therefore, it is pertinent to locate well the path to be followed by the end-effector of a robot within its workspace in order to minimize the energy consumed by its actuators. Accordingly, the major contribution of this paper are \textit{i})~the minimization of the energy requirements by optimum path placement and \textit{ii})~use of electric energy consumed by the actuators instead of treating mechanical energy relations.  Hence, we propose an approach to optimize the location of a given path within the workspace of a manipulator in order to minimize the electric energy consumed by its actuators.
It should be noted that energy optimal poses can affect various performance indices such as manipulability, dexterity, stiffness etc. In the scope of this paper, we have only considered the electric energy consumption as an optimization criterion (objective function) in order to highlight its influence on motion planning. However, others optimization criteria , such as the manipulability, dexterity, stiffness, the motors torque required, will be taken into account in future work. Accordingly, we will come up with multiobjective path placement optimization problem that may not be convex. For that matter, we will use other optimization tools such as Genetic Algorithms to solve it.

The paper is organized as follows. In section II, we propose a minimum energy path placement optimization problem that includes basic problem formulation, electric energy calculations and the algorithm proposed to solve the problem. In section III, the Orthoglide 3-axis: a three-degree-of-freedom translational Parallel Kinematics Machine (PKM), is used as a case study to implement the proposed optimization methodology and results are presented for rectangular test paths.
\section{PATH PLACEMENT OPTIMIZATION}
The problem aims at determining the optimal location of a predefined path in order to minimize the electric energy used by the actuators. The optimization problem is subject to geometric, kinematics and dynamics constraints. Geometric constraints include joint limits and the boundaries of the workspace. Kinematic constraints deal with the maximum actuator velocities whereas dynamic constraints are related to actuator wrenches. Contrary to the trajectories usually defined with start and end point configurations, the entire path is supposed to be known within the framework of this research work. The path location can be defined in a similar way as to define the location of a workpiece with respect to a manipulator reference point.
\subsection{Path Localization}
\begin{figure}[h]			
\centering
\begin{tabular}{cc}
  \psfrag{xb}[c][c][0.8]{$X_b$}
  \psfrag{yb}[c][c][0.8]{$Y_b$}
  \psfrag{zb}[c][c][0.8]{$Z_b$}
  \psfrag{Ob}[c][c][0.8]{$O_b$}
  \psfrag{Fb}[c][c][0.8]{${\cal F}_b$}
  \psfrag{xp}[c][c][0.7]{\textcolor{myred}{$X_p$}}
  \psfrag{yp}[c][c][0.7]{\textcolor{myred}{$Y_p$}}
  \psfrag{zp}[c][c][0.7]{\textcolor{myred}{$Z_p$}}
  \psfrag{Op}[c][c][0.7]{\textcolor{myred}{$O_p$}}
  \psfrag{Fp}[c][c][0.7]{\textcolor{myred}{${\cal F}_p$}}
  \psfrag{T}[c][c][0.7]{\textcolor{myred}{${\cal P}$}}
  \psfrag{pv}[c][c][0.7]{\textcolor{myred}{$\mathbf{p_{{\cal F}_p}}$}}
  \psfrag{P}[c][c][0.8]{\textcolor{myblue}{$P$}}
  \psfrag{a}[c][c][0.8]{($a$)}
\epsfig{file=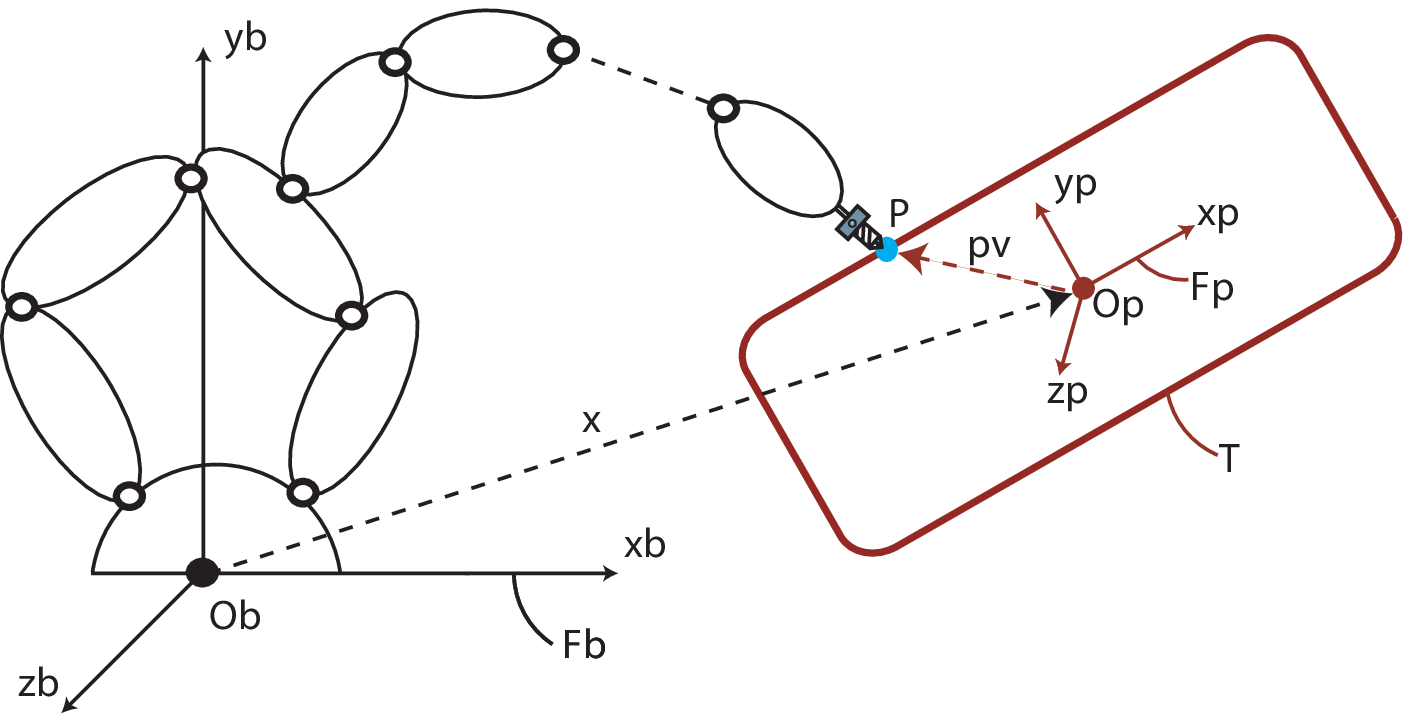,width=0.65\linewidth} &
  \psfrag{Xb}[c][c][0.8]{$X_b$}
  \psfrag{Yb}[c][c][0.8]{$Y_b$}
  \psfrag{Zb}[c][c][0.8]{$Z_b$}
  \psfrag{O}[c][c][0.8]{$O$}
  \psfrag{Xp}[c][c][0.7]{$X_p$}
  \psfrag{Yp}[c][c][0.7]{$Y_p$}
  \psfrag{Zp}[c][c][0.7]{$Z_p$}
  \psfrag{u}[c][c][0.9]{$u$}
  \psfrag{v}[c][c][0.9]{$v$}
  \psfrag{w}[c][c][0.9]{$w$}
  \psfrag{phi}[c][c][0.7]{$\phi$}
  \psfrag{si}[c][c][0.7]{$\psi$}
  \psfrag{th}[c][c][0.7]{$\theta$}
  \psfrag{phid}[c][c][0.8]{$\dot{\phi}$}
  \psfrag{sid}[c][c][0.8]{$\dot{\psi}$}
  \psfrag{thd}[c][c][0.8]{$\dot{\theta}$} 
  \psfrag{b}[c][c][0.8]{($b$)}
\epsfig{file=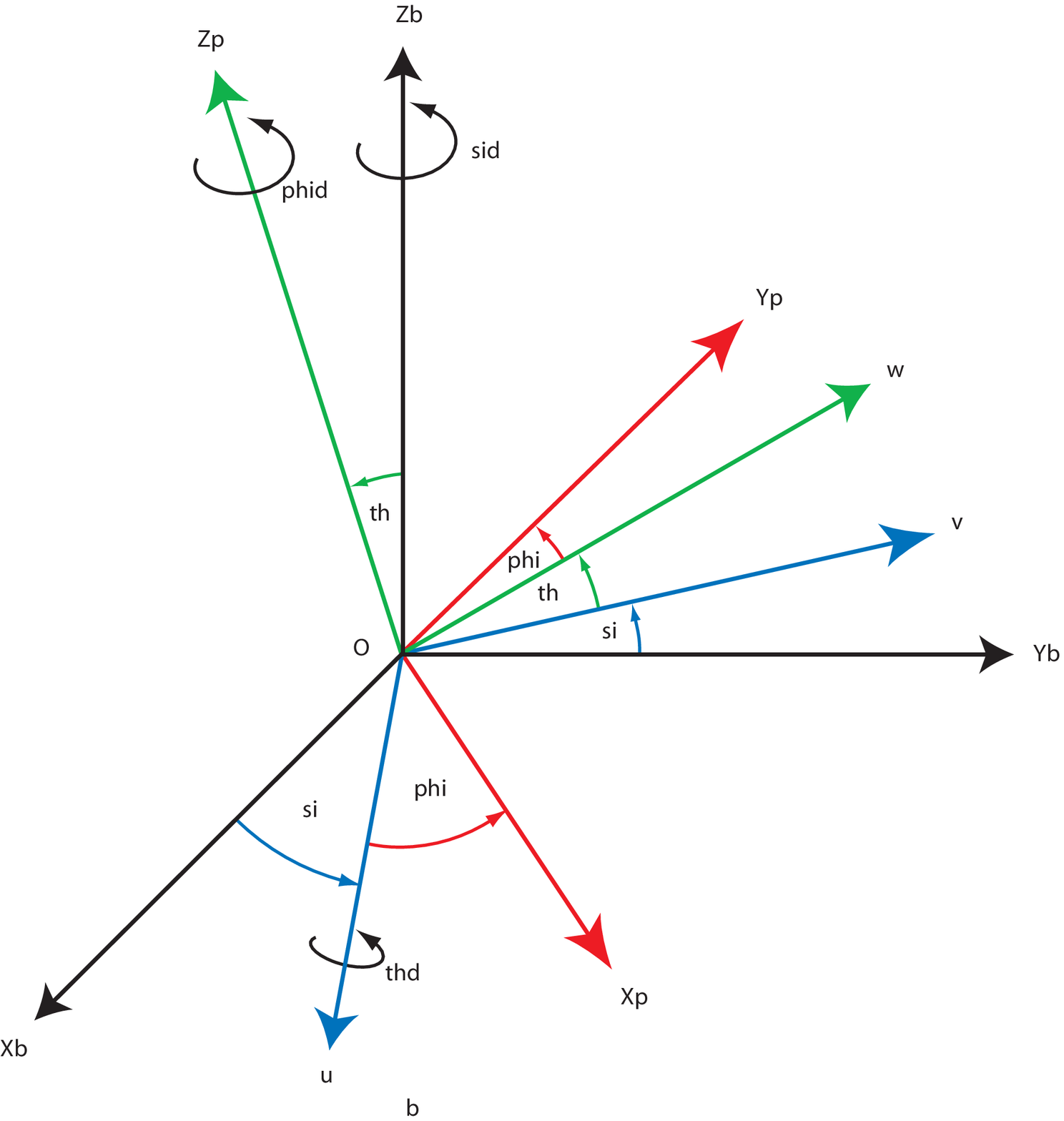,width=0.35\linewidth,angle=0}
\end{tabular}
\caption{(a). Path ${\cal P}$ to be followed by the end-effector $P$ of a manipulator, ${\cal F}_b$ and ${\cal F}_p$ being the base and path frames, (b). Euler angles}
\label{fig:FbFpFrames}
\end{figure}

In order to formulate and describe the problem, two reference frames are defined: $i$)~the path frame ${\cal F}_p$ and $ii$)~the base frame ${\cal F}_b$, as shown in Fig.~\ref{fig:FbFpFrames}. The path frame ${\cal F}_p$, is attached to the given/required path at a suitable point such as geometric center of the path. As ${\cal F}_p$ is attached to the path, the end-effector trajectory parameters remains constant in this reference frame, no matter where it is located. In other words, the path is fully defined and constant in ${\cal F}_p$. It can also be named workpiece frame as it characterizes the position and the orientation of the workpiece within the manipulator workspace. 
The base frame ${\cal F}_b$ can also be called global or manipulator frame. It is attached to the manipulator base and is used to locate a workpiece (or ${\cal F}_p$) with respect to the manipulator coordinate system. The location and orientation of ${\cal F}_p$ with respect to ${\cal F}_b$ can be defined in such a way that the whole path lies within the workspace.
The position of ${\cal F}_p$ with respect to ${\cal F}_b$ is defined with the Cartesian coordinates of the origin of ${\cal F}_p$ and the relative orientation of the two frames is characterized by means of Euler angles. 
However, keeping in view the constraints of the manipulator wrist, Euler angles are uniquely defined in the context of milling operation with the parameterization given in Fig.~\ref{fig:FbFpFrames}(b). It allows to avoid the singularity of Euler parameters.
   As a matter of fact, any trajectory defined in ${\cal F}_p$ can be transformed in the base frame ${\cal F}_b$ by a transformation matrix. For instance, point $P$, of Cartesian coordinates $x_{P_p},y_{P_p},z_{P_p}$ in ${\cal F}_p$ can be expressed in ${\cal F}_b$ as follows:
\begin{equation}
\left[ \begin{array}{c}  
			\mathbf{p}
\end{array}
\right]_{{\cal F}_b}
=\mathbf{^\textit{b}T_\textit{p}}
\left[ \begin{array}{cc}
			\mathbf{p}
\end{array}
\right]_{{\cal F}_p}
\label{Eq:Fp2Fb1}
\end{equation}
namely,
\begin{equation}
\left[	\begin{array}{cccc}
				x_{P_b}	&	y_{P_b}	&		z_{P_b}	&		1 \end{array}	\right]^T_{{\cal F}_b}
=\mathbf{^\textit{b}T_\textit{p}}
\left[	\begin{array}{cccc}
				x_{P_p}		&		y_{P_p}		&		z_{P_p}		&	1 \end{array}	\right]^T_{{\cal F}_p}
\label{Eq:Fp2Fb}
\end{equation}
\noindent $\mathbf{^\textit{b}T_\textit{p}}$ being the transformation matrix from ${\cal F}_p$ to ${\cal F}_b$. Let $O_p(x_{O_p} , y_{O_p} ,z_{O_p})$ be the origin of the path frame expressed in frame ${\cal F}_b$ and $(\phi , \theta , \psi)$ the Euler angles characterizing the orientation of frame ${\cal F}_p$ with respect to frame ${\cal F}_b$. Accordingly,
  \begin{equation}
  \mathbf{^\textit{b}T_\textit{p}}= \left[
  							\begin{array}{cccc}
  							\cos{\phi}\cos{\theta}	&	\cos{\phi}\sin{\theta}\sin{\psi}-\sin{\phi}\cos{\psi}& \cos{\phi}\sin{\theta}\cos{\psi}+\sin{\phi}\sin{\psi}	&	x_{O_p}		\\
  							\sin{\phi}\cos{\theta}	&	\sin{\phi}\sin{\theta}\sin{\psi}+\cos{\phi}\cos{\psi}& \sin{\phi}\sin{\theta}\cos{\psi}-\cos{\phi}\sin{\psi}	&	y_{O_p}		\\
  							-\sin{\theta}		&		\cos{\theta}\sin{\psi}	&	 \cos{\theta}\cos{\psi}	&	z_{O_p}		\\
  							0		&		0	&	 0	&	1		\\
  							\end{array}
  							\right]
  \label{Eq:bTp}
	\end{equation}
The path placement is specified with  $\mathbf{^\textit{b}T_\textit{p}}$. Let $\mathbf{x}=[x_{O_p} \quad y_{O_p} \quad z_{O_p} \quad \phi \quad \theta \quad \psi]^T$
define the path placement within the workspace of the manipulator, in the reference frame ${\cal F}_b$. Consequently, the components of $\mathbf{x}$ are the decision variables of the optimization problem at hand.\\
In the context of a general machining process like milling operation, the feature to be machined on the workpiece is defined with respect to the frame attached to the workpiece, namely ${\cal F}_p$. Likewise, the machining operation conditions such as machining velocity and acceleration are fully defined in ${\cal F}_p$. Finally, the part to be machined is defined by the designer and located in $\mathcal{F}_b$ whereas the machining operation conditions and robot trajectory planning are defined by the production engineer with respect to the the corresponding part, namely $\mathcal{F}_p$. Here, we introduce a methodology to help the production engineer well locate the workpiece, namely $\mathcal{F}_p$, within the robot base frame $\mathcal{F}_b$ in order to minimize the actuators electric energy consumption.
\subsection{Path Placement Optimization Problem Formulation}
\label{Sec:OPF}
The goal of this research work is to help the path planner find the best location of the path to be followed by the robot in order to minimize the energy used by its actuators. It can be formulated as an optimization problem, namely,\\
``\emph{For a predefined path in ${\cal F}_p$, find the optimum location and orientation of ${\cal F}_p$ with respect to ${\cal F}_b$, defined by the decision variables $\mathbf{x}$, in order to minimize the electric energy used by the manipulator actuators to generate that path, while respecting the geometric, kinematic and dynamic constraints of the manipulator.}''
It can also be formulated mathematically as follows:
\begin{equation}		
\min_{\mathbf{x}} E_t=\sum^n_{i=1}{E_i}\left( \mathbf{x} \right) \qquad
\textrm{ subject to:} \left\{ 
									\begin{array}{l l}
  								q_{il}\leq q_i\leq q_{iu}\\
									\left| \dot{q}_i \right| \leq \dot{q}_{iu}   \qquad \qquad (i=1~\cdots~n)\\		
									\left| \tau_i \right| \leq \tau_{iu}	 
									\end{array} \right. 
\label{Eq:ProbStm}
\end{equation}	
$\mathbf{x}$ is the path placement vector corresponding to the transformation matrix $\mathbf{^\textit{b}T_\textit{p}}$. $E_t$ is the total electric energy required by the $n$ actuators whereas $E_i$ is the total electric energy required by the $i^{th}$ actuator to follow the path.
$q_i$, $\dot{q}_i$, $\tau_i$ are respectively the $i^{th}$ actuator displacement, rate and torque. $q_{il}$ is the lower bound and $q_{iu}$ (resp. $\dot{q}_{iu}$ and  $\tau_{iu}$) is the upper bound of $i^{th}$ actuator displacement (resp. rate and torque). For a given path placement vector $\mathbf{x}$, these constraints can be evaluated by means of the manipulator kinematic, velocity and dynamic models.\\
It is noteworthy that the manipulator geometric constraints guarantee that the whole path lies inside the prescribed workspace. Similarly, the bounds on actuator rates ($\dot{q}_{iu}$) and torques ($\tau_{iu}$) ensure that the manipulator will not go through any singular configuration while following the path. The electric energy used by the actuators is formulated in the next section.
\subsection{Objective Function: Electric Energy}
\label{Sec:ElectEnergy}
The energy used by the motors depends on their corresponding velocities and torques. As a matter of fact, the electric current in the motors varies with motor velocities and torques. Accordingly, the motor's self-inductance phenomenon appears. The current $I$ drawn by the motors and the motor electromotive potential $V_e$ can be calculated as a function of the required torque $\tau$ and the angular velocity $\omega$ of the actuators, namely,
		\begin{equation}
		\label{Equ:current}
					I=\frac{\tau}{K_t}
				\label{Eq:i}
		\end{equation}
		\begin{equation}
		\label{Equ:voltage}
					V_e=K_e\omega 
				\label{Eq:Ve}
		\end{equation}
$K_t$ being the torque sensitivity factor or motor constant expressed in [Nm/A] and $K_e$ the back electromotive force constant expressed in [V.(rad/sec)$^{-1}$].\\
The total electric power $P_T$ is composed of \cite{GLacroux1994}:
\begin{itemize}
	\item The resistive power loss (Joule effect):
 				\begin{equation}
				P_J=RI^2 
				\label{Eq:Pj}
			\end{equation}
	\item the inductive power loss:
 				\begin{equation}
				P_L=LI\frac{dI}{dt}
				\label{Eq:Pl}
			\end{equation}
 	\item the power used to produce the electromotive force:
 				\begin{equation}
				P_{EM}=V_eI
				\label{Eq:Pem}
			\end{equation}
 	\end{itemize}
Accordingly, the total electric power $P_T$ can be expressed as follows:
	\begin{equation}
		P_T=P_J+P_L+P_{EM} 
		\label{Eq:Pe}
	\end{equation}
\textit{R} being the motor winding resistance expressed in Ohm [$\Omega$] and \textit{L} the motor inductance coefficient expressed in Henry [H].\\
Finally, the energy \textit{E} consumed by a motor can be evaluated by integrating $P_T$ over the total trajectory time \textit{T}, namely,
\begin{equation}
	E = \int_0^T P_T dt
	\label{Eq:E}
\end{equation}
$P_T$ being the instantaneous electric power at time \textit{t}, defined in Eq.~(\ref{Eq:Pe}).\\
It should be noted that Eq.~(\ref{Eq:i}) allows us to consider the energy used by the actuators when they do not move but still produce a torque to keep the manipulator at a certain stationary configuration (with respect to that particular direction or actuator), like resisting the gravity.

It is noteworthy that energy calculation model presented in this section is suitable for the brushless motors, which are generally used for PKMs. However, depending on the motors/derives in application, energy calculation model can be developed accordingly.
\subsection{Resolution of the Path Placement Optimization Problem}
\begin{figure}[tbp]
  \centering
  \psfrag{A}[c][c][0.8]{\textcolor{myblue}{Preparation Phase}}
  \psfrag{A11}[c][c][0.8]{Geometric \& Dynamic Parameters}
  \psfrag{A12}[c][c][0.8]{Definition of base frame ${\cal F}_b$}
  \psfrag{A13}[c][c][0.8]{Electric motors parameters}
  \psfrag{A14}[c][c][0.8]{Path definition}
  \psfrag{A2}[c][c][0.8]{Definition of Path frame ${\cal F}_p$}
  \psfrag{A3}[c][c][0.8]{Trajectory Definition in ${\cal F}_p$}
  \psfrag{A4}[c][c][0.8]{Initial Guess $\mathbf{x}_0$}
  \psfrag{A41}[c][c][0.8]{$\mathbf{x}_{_0}=[x_{{Op}_0} \quad y_{{Op}_0} \quad z_{{Op}_0} \quad \phi_{_0} \quad \theta_{_0} \quad \psi_{_0} ]^T$}

  \psfrag{B}[c][c][0.8]{\textcolor{myblue}{Evaluation}}			
  \psfrag{BB}[c][c][0.8]{\textcolor{myblue}{Phase}}	
  \psfrag{BBB}[c][c][0.6]{\textcolor{myblue}{}}	
  \psfrag{B1}[c][c][0.8]{Transformation Matrix $\mathbf{^\textit{b}T_\textit{p}}$}
  \psfrag{B2}[c][c][0.8]{Trajectory Definition in ${\cal F}_b$}
  \psfrag{B31}[c][c][0.8]{Inverse Geometric Model}
  \psfrag{B32}[c][c][0.8]{Inverse Kinematic Model}
  \psfrag{B33}[c][c][0.8]{Inverse Dynamic Model}
  \psfrag{B4}[c][c][0.8]{Objective and Constraints Evaluation}
  
  \psfrag{C}[c][c][1]{\textcolor{myblue}{Optimization Phase}}
  \psfrag{C1}[c][c][0.8]{$q_{il}\leq q_i\leq q_{iu}$}
  \psfrag{C12}[c][c][0.8]{$\left| \dot{q}_i \right| \leq \dot{q}_{iu}$}
  \psfrag{C13}[c][c][0.8]{$\left| \tau_i \right| \leq \tau_{iu}$}
  
  \psfrag{C2}[c][c][1]{$E_t=E_{min}$}
  \psfrag{C3}[c][c][1]{$\mathbf{x}_j=\mathbf{x}_{j+1}$}
  
  \psfrag{D1}[c][c][1]{Optimum value}
  \psfrag{D2}[c][c][1]{$\mathbf{x}^*=\mathbf{x}_j$}
  
  \psfrag{Y1}[c][c][0.8]{Yes}
  \psfrag{Y2}[c][c][0.8]{Yes}
  \psfrag{N1}[c][c][0.8]{No}
  \psfrag{N2}[c][c][0.8]{No}

  \psfrag{MGI}[c][b][0.8]{IKM, IVM and IDM}

	\includegraphics[width=14cm]{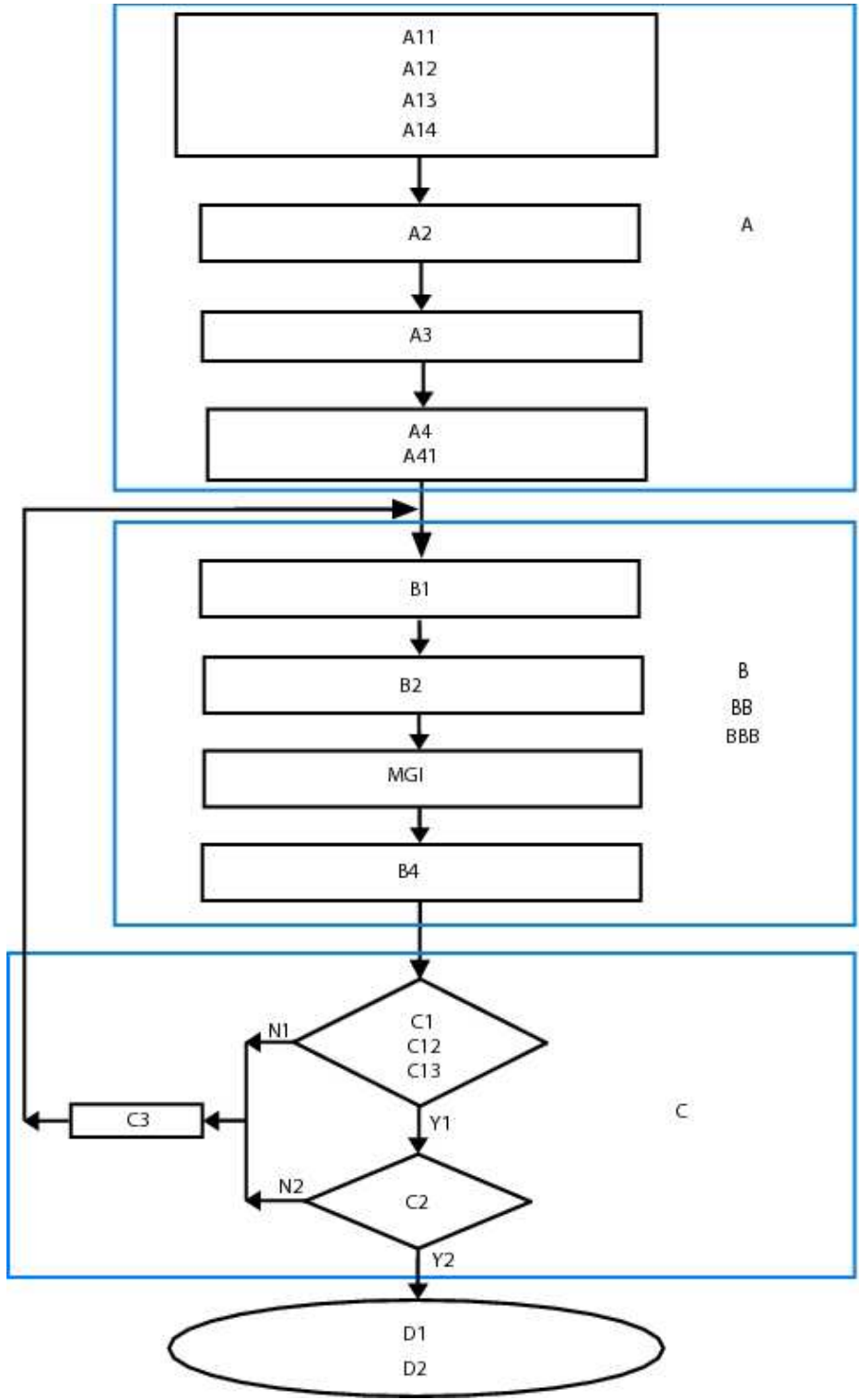}
	\caption{Flowchart of the path placement optimization process}
  \label{fig:Flowchart}
	\end{figure}

To solve the problem, a general optimization approach is proposed as illustrated in Fig.~\ref{fig:Flowchart}. The approach can be summarized in three constituent elements or phases:
\begin{enumerate}
	\item 
	Preparation Phase:	Manipulator geometric, dynamic and electric parameters along with the definition of the required path are used as the known input data of the optimization problem. The base frame ${\cal F}_b$ is defined. The path to be followed by the end-effector of the robot is defined in the path frame $\mathcal{F}_p$. The terms of the transformation matrix between $\mathcal{F}_p$ and $\mathcal{F}_b$ are the decision variables of the optimization problem.
	\item
	Evaluation Phase: At this stage, the inverse kinematic  model (IKM), the inverse velocity model (IVM) and the inverse dynamic model (IDM) of the manipulator are determined for each set of design parameters obtained from the optimization routine. Accordingly, the objective function and the constraints of the optimization problem are evaluated. 
	\item
	Optimization Phase: The objective function and constraints evaluated at the previous step are analyzed by means of the optimization algorithm. Once the constraints are satisfied, the objective function is tested for its optimum value. The convergence criteria are checked. If latter are respected, the optimization algorithm stops. Otherwise, other iterations are run as long as the convergence criteria are not satisfied.
	\end{enumerate}
Finally, the optimum path placement is obtained by means of the position and orientation of the origin of ${\cal F}_p$ with respect to ${\cal F}_b$ defined by the terms of the transformation matrix between ${\cal F}_p$ and ${\cal F}_b$, namely, the location vector $\mathbf{x}^*$=[$x_{O_p}^* \quad y_{O_p}^* \quad  z_{O_p}^* \quad   \phi^* \quad   \theta^* \quad  \psi^* $]$^T$.\
\section{CASE STUDY:\\ APPLICATION TO THE ORHTOGLIDE}
\subsection{Description of the Orthoglide}
\begin{figure}[ht]
\centering
\subfigure[Orthoglide (courtesy:\,CNRS Phototh\'{e}que/CARLSON Leif)]{
\includegraphics[height=6.5cm]{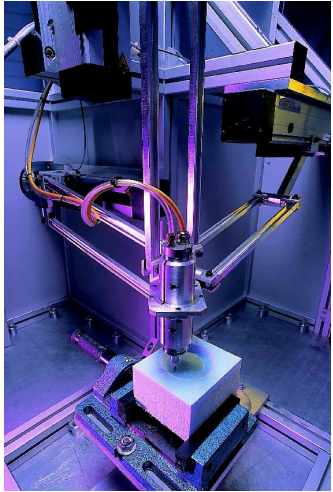}
\label{fig:Ortho3axis}
} 
\hspace{1cm}
\subfigure[Cubic workspace ($200\times200\times200$ mm$^{3}$)]{
		\psfrag{s01}[cc][lt][0.8]{\color[rgb]{0,0,0}\setlength{\tabcolsep}{0pt}\begin{tabular}{l}$X$ [mm]\end{tabular}}%
		\psfrag{s02}[cc][lt][0.8]{\color[rgb]{0,0,0}\setlength{\tabcolsep}{0pt}\begin{tabular}{r}$Y$ [mm]\end{tabular}}%
		\psfrag{s03}[b][b][0.8]{\color[rgb]{0,0,0}\setlength{\tabcolsep}{0pt}\begin{tabular}{c}$Z$ [mm]\end{tabular}}%
		\psfrag{s04}[r][r][0.8]{\textcolor{green}{$C$}}%
		\psfrag{s05}[r][r][0.8]{\textcolor{blue}{$O_b$}}%
		\psfrag{s06}[r][r][0.8]{\textcolor{red}{$Q^+$}}%
		\psfrag{s07}[l][l][0.8]{\textcolor{red}{$Q^-$}}%
		\psfrag{s08}[c][c][0.8]{\textcolor{blue}{$X_b$}}%
		\psfrag{s09}[c][c][0.8]{\textcolor{blue}{$Y_b$}}%
		\psfrag{s10}[r][r][0.8]{\textcolor{blue}{$Z_b$}}%
		\psfrag{x01}[t][t][0.8]{0}					\psfrag{x02}[t][t][0.8]{0.1}%
		\psfrag{x03}[t][t][0.8]{0.2}				\psfrag{x04}[t][t][0.8]{0.3}%
		\psfrag{x05}[t][t][0.8]{0.4}				\psfrag{x06}[t][t][0.8]{0.5}%
		\psfrag{x07}[t][t][0.8]{0.6}				\psfrag{x08}[t][t][0.8]{0.7}%
		\psfrag{x09}[t][t][0.8]{0.8}				\psfrag{x10}[t][t][0.8]{0.9}%
		\psfrag{x11}[t][t][0.8]{1}					\psfrag{x12}[t][t][0.8]{-100}%
		\psfrag{x13}[t][t][0.8]{-50}				\psfrag{x14}[t][t][0.8]{0}			\psfrag{x15}[t][t][0.8]{50}%
		\psfrag{v01}[r][r][0.8]{0}					\psfrag{v02}[r][r][0.8]{0.1}%
		\psfrag{v03}[r][r][0.8]{0.2}				\psfrag{v04}[r][r][0.8]{0.3}%
		\psfrag{v05}[r][r][0.8]{0.4}				\psfrag{v06}[r][r][0.8]{0.5}%
		\psfrag{v07}[r][r][0.8]{0.6}				\psfrag{v08}[r][r][0.8]{0.7}%
		\psfrag{v09}[r][r][0.8]{0.8}				\psfrag{v10}[r][r][0.8]{0.9}%
		\psfrag{v11}[r][r][0.8]{1}					\psfrag{v12}[r][r][0.8]{-100}%
		\psfrag{v13}[r][r][0.8]{-50}				\psfrag{v14}[r][r][0.8]{0}			\psfrag{v15}[r][r][0.8]{50}%
		\psfrag{z01}[r][r][0.75]{-120}			\psfrag{z02}[r][r][0.75]{-100}%
		\psfrag{z03}[r][r][0.75]{-80}				\psfrag{z04}[r][r][0.75]{-60}%
		\psfrag{z05}[r][r][0.75]{-40}				\psfrag{z06}[r][r][0.75]{-20}%
		\psfrag{z07}[r][r][0.75]{0}					\psfrag{z08}[r][r][0.75]{20}%
		\psfrag{z09}[r][r][0.75]{40}				
\includegraphics[width=8cm,height=6cm]{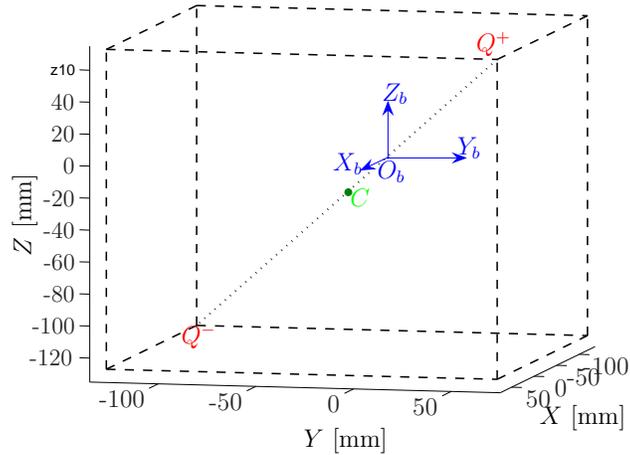}
\label{fig:OrthoWS}
}
\caption[Optional caption for list of figures]{A snap and workspace of the Orthoglide}
\label{fig:Orthoglide}
\end{figure}

The Orthoglide is a Delta-type PKM \cite{Clavel1988} dedicated to 3-axis rapid machining applications developed in IRCCyN \cite{Wenger2000}. It gathers the advantages of both serial and parallel kinematic architectures such as regular workspace, homogeneous performances, good dynamic performances and stiffness. The Orthoglide is composed of three identical legs, as shown in Fig.~\ref{fig:Ortho3axis}. Each leg is made up of a prismatic joint, two revolute joints and a parallelogram joint.
Prismatic joints of the legs, mounted orthogonally, are actuated which result the motion of the mobile platform in the Cartesian space with fixed orientation.\\
\begin{table}[htbp]
	\centering
		\begin{tabular}{cc}
		\hline
			\multicolumn{2}{c}{Workspace size $L_{workspace}=0.2$ m}\\
		\hline
			Point	&	Cartesian coordinates in ${\cal F}_b$ [m]\\
		\hline
		$O_b$ & $(0, 0, 0)$\\
		\hline
		$C$ & $(-0.027, -0.027, -0.027)$\\
		\hline
		$Q^+$ & $(0.73, 0.73, 0.73)$\\
		\hline
		$Q^-$ & $(-0.127, -0. 127, -0.127)$\\
		\hline		
		\end{tabular}
	\caption{Orthoglide workspace parameters}
	\label{Tab:OrthoWS}
\end{table}
The Orthoglide 3-axis geometric parameters are function of the size of the prescribed cubic Cartesian workspace, defined by the length of the cube sides, namely, $L_{workspace}$ \cite{Pashkevich2007}. 
The base frame ${\cal F}_b$ is defined with the unit vector $\mathbf{e}_i$ in the direction of the $i^{th}$ prismatic joint, namely, $X_b$, $Y_b$ and $Z_b$, the origin $O_b$ of ${\cal F}_b$ being the intersecting point of $\mathbf{e}_i$. Two points $Q^+$ and $Q^-$ are defined in such a way that the velocity transmission factor is $1/2$ and $2$ at these two points \cite{Chablat2003}. A cube is then constructed with $Q^+ Q^-$ as its diagonal. It should be noted that the cubic workspace center, i.e., point $C$, and the origin $O_b$ of the reference frame ${\cal F}_b$ do not coincide, as shown in Fig.~\ref{fig:OrthoWS}. In the scope of this study, $L_{workspace}$ is equal to $0.200$m. Accordingly, the coordinates of points $Q^+$, $Q^-$ and $C$ for the said workspace are  given in Table~\ref{Tab:OrthoWS}. Similarly, the prismatic actuator bounds, $\rho_{min}$ and $\rho_{max}$, can be calculated \cite{Pashkevich2007}. Table~\ref{Tab:OrthoConstr} shows the lower and upper bounds of the prismatic joints displacements and their maximum allowable velocity and torque for the Orthoglide. The geometric, kinematic and dynamic parameters of the Orthoglide are defined in \cite{Wenger2000,Pashkevich2007,Chablat2003, Guegan2003}.\\
Electric energy $E_i$ used by each actuator is calculated by means of Eqs.~(\ref{Eq:i}) to (\ref{Eq:E}). As the Orthoglide~3-axis has three \textit{3-phase Sanyo Denki} synchronous servo motors ($reference: P30B0604D$), Eq.~(\ref{Eq:Pe}) is multiplied by $3$ to cater for the power consumed by the each phase of the motor in order to calculate the electric power $P_{Ti}$ used by each actuator, i.e.,
	\begin{equation}
	\label{Eq:PowerOrtho}
		P_{T_i}=3(RI^2+LI\frac{dI}{dt}+V_eI)
	\end{equation}
	\begin{table}[htbp]
	\centering
		\begin{tabular}{cl}
		\hline
		$\quad\rho_{i_{min}}\quad\quad$	&	 \quad $0.126$\,m \quad      \\
		\hline
		$\quad\rho_{i_{max}}\quad\quad$ &  \quad $0.383$\,m	\quad		\\
		\hline
		$\quad v_{i_{max}}\quad\quad$    & \quad $1.00$\,m.s$^{-1}$\quad	\\
		\hline
		$\quad\tau_{i_{max}}\quad\quad$ &  \quad $ 1.274$\,Nm\quad			\\
		\hline
		\end{tabular}
	\caption{Orthoglide actuators parameters ($i=x,y,z$)}
	\label{Tab:OrthoConstr}
\end{table}

\subsection{Trajectory Planning and External Forces}
\begin{figure}[h]			
\centering
\begin{tabular}{cc}
  \psfrag{xxb}[c][c][0.8]{$X_b$}
  \psfrag{yyb}[c][c][0.8]{$Y_b$}
  \psfrag{Ob}[c][c][0.8]{$O_b$}
  \psfrag{pFb}[c][c][0.8]{$\mathbf{p_{{\cal F}_b}}$}
  \psfrag{Fb}[c][c][0.8]{${\cal F}_b$}
  \psfrag{Xp}[c][c][0.8]{\textcolor{myblue}{$X_p$}}
  \psfrag{Yp}[c][c][0.8]{\textcolor{myblue}{$Y_p$}}
  \psfrag{Op}[c][c][0.8]{\textcolor{myblue}{$O_p$}}
  \psfrag{P}[c][c][0.8]{\textcolor{myblue}{$P$}}
  \psfrag{pFp}[c][c][0.8]{\textcolor{myblue}{$\mathbf{p_{{\cal F}_p}}$}}
  \psfrag{Fp}[c][c][0.8]{\textcolor{myblue}{${\cal F}_p$}}
  \psfrag{A}[c][c][0.7]{$A$}
  \psfrag{B}[c][c][0.7]{$B$}
  \psfrag{C}[c][c][0.7]{$C$}
  \psfrag{D}[c][c][0.7]{$D$}
  \psfrag{L}[c][c][0.7]{$L$}
  \psfrag{W}[c][c][0.7]{$W$}
  \psfrag{Phi}[c][c][0.9]{$\phi$}
  \psfrag{Ppp}[c][c][1]{P}
  \psfrag{Ppp}[c][r][0.8]{\textcolor{myblue}{$Path~{\cal P}$}}
  \psfrag{Xb}[c][c][1]{$X_b$~[mm]}
  \psfrag{Yb}[c][c][1]{$Y_b$~[mm]}
  \psfrag{a}[c][c][1]{($a$)}
\epsfig{file=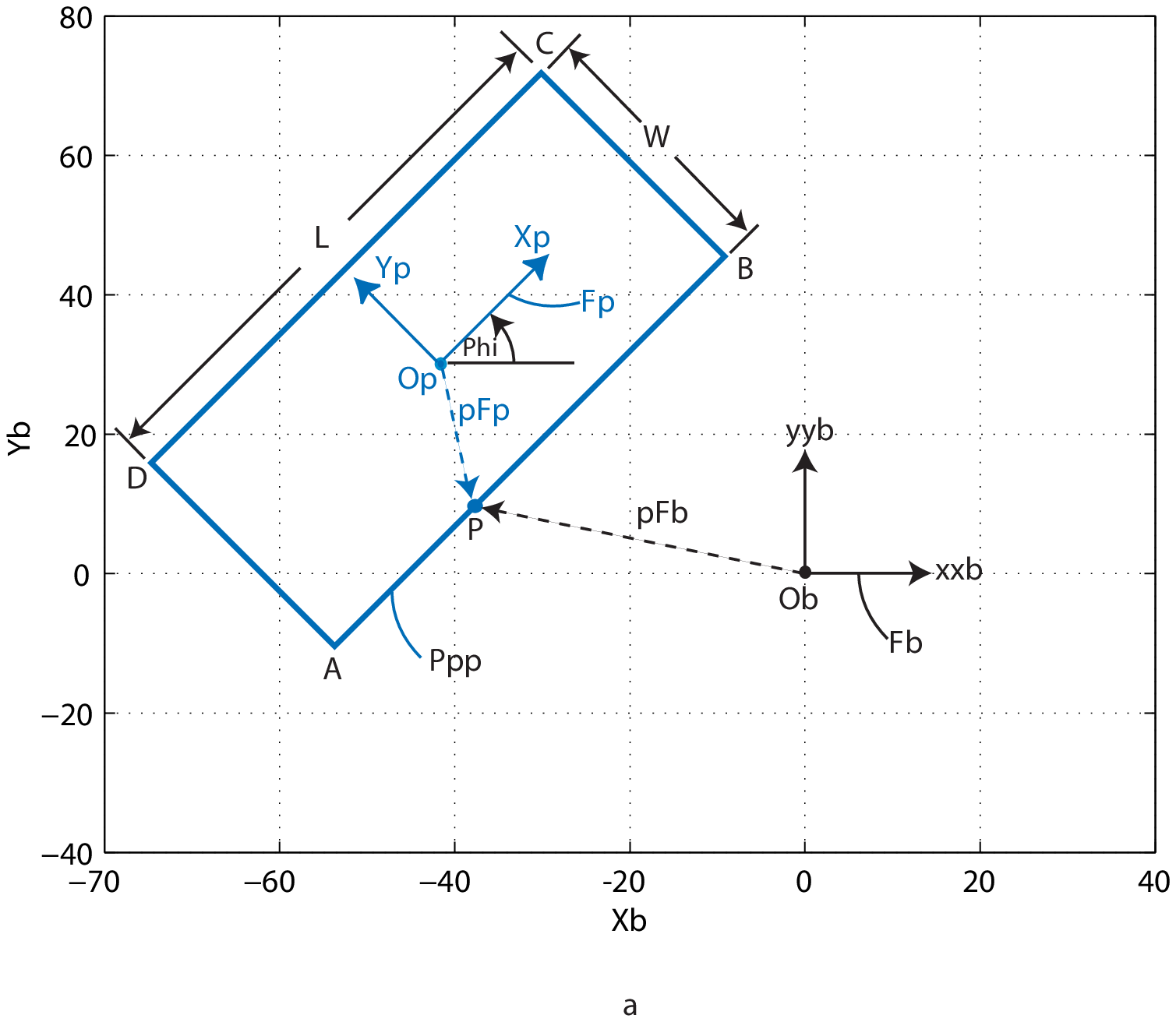,width=0.7\linewidth} &
  \psfrag{b}[c][c][1]{($b$)}
  \psfrag{WP}[c][c][0.7]{Workpiece}
  \psfrag{RP}[c][c][0.7]{Rectangular Pocket}
  
\epsfig{file=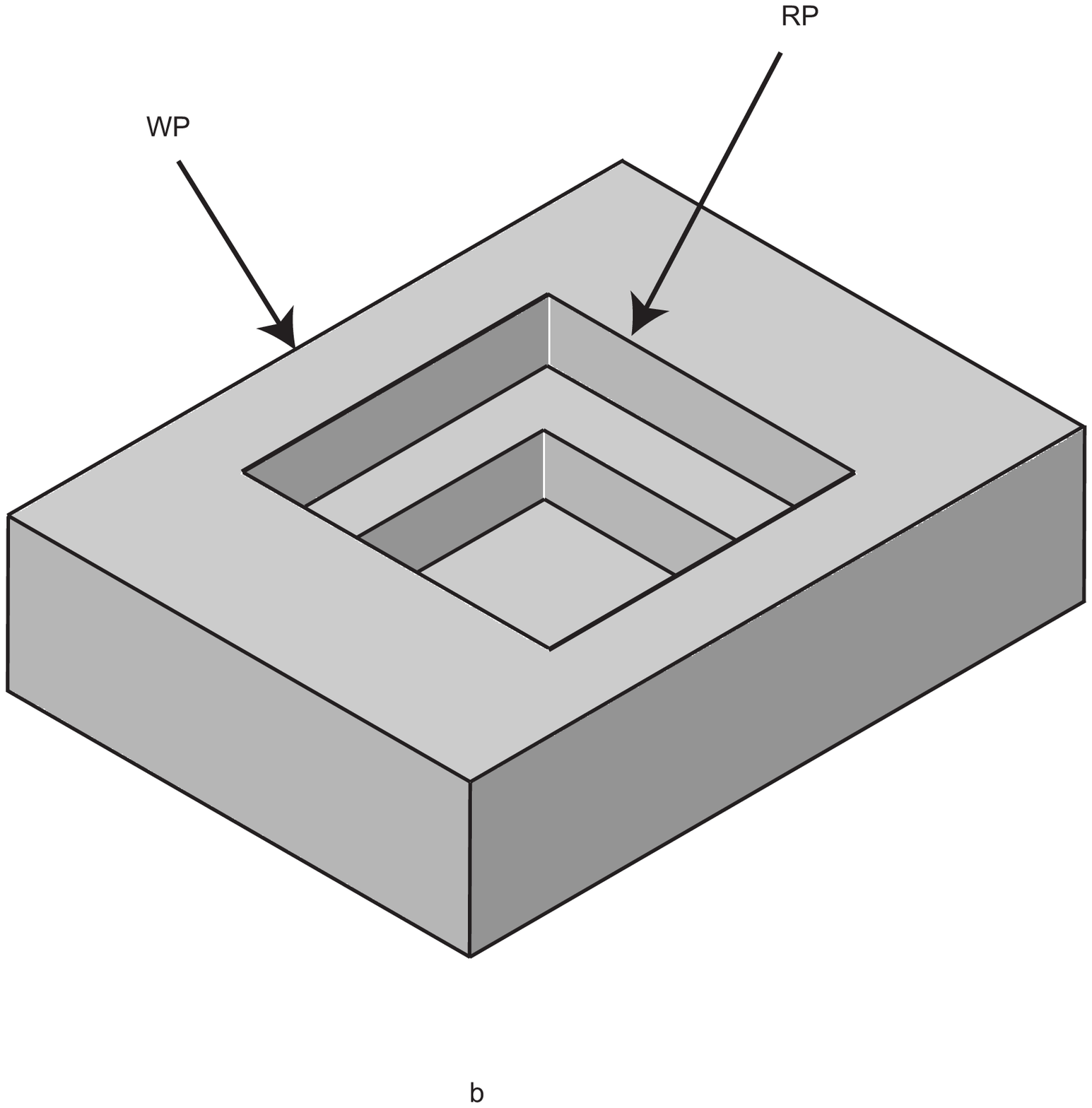,width=0.3\linewidth,angle=0} 
\end{tabular}
	 \caption{Rectangular test path}
   \label{fig:TestPath}
\end{figure}

\begin{figure}[h]		
  \centering
		\psfrag{s10}[l][l][0.8]{[m]}%
		\psfrag{s12}[c][c][0.9]{End-effector position vs time in {${\cal F}_p$}}%
		\psfrag{s14}[l][l][0.8]{Time [sec]}%
		\psfrag{s15}[l][l][0.8]{[m.s$^{-1}$]}%
		\psfrag{s17}[c][c][0.9]{End-effector velocity vs time in {${\cal F}_p$}}%
		\psfrag{s19}[l][l][0.7]{$z_{P_p}$}%
		\psfrag{s20}[l][l][0.7]{$x_{P_p}$}%
		\psfrag{s21}[l][l][0.7]{$y_{P_p}$}%
		\psfrag{s22}[l][l][0.7]{$z_{P_p}$}%
		\psfrag{s23}[l][l][0.7]{$\dot{z}_{P_p}$}%
		\psfrag{s24}[l][l][0.7]{$\dot{x}_{P_p}$}%
		\psfrag{s25}[l][l][0.7]{$\dot{y}_{P_p}$}%
		\psfrag{s26}[l][l][0.7]{$\dot{z}_{P_p}$}%
		\psfrag{x01}[t][t][0.8]{0}%
		\psfrag{x02}[t][t][0.8]{0.1}%
		\psfrag{x03}[t][t][0.8]{0.2}%
		\psfrag{x04}[t][t][0.8]{0.3}%
		\psfrag{x05}[t][t][0.8]{0.4}%
		\psfrag{x06}[t][t][0.8]{0.5}%
		\psfrag{x07}[t][t][0.8]{0.6}%
		\psfrag{x08}[t][t][0.8]{0.7}%
		\psfrag{x09}[t][t][0.8]{0.8}%
		\psfrag{x10}[t][t][0.8]{0.9}%
		\psfrag{x11}[t][t][0.8]{1}%
		\psfrag{x12}[t][t][0.8]{0}%
		\psfrag{x13}[t][t][0.8]{0.1}%
		\psfrag{x14}[t][t][0.8]{0.2}%
		\psfrag{x15}[t][t][0.8]{0.3}%
		\psfrag{x16}[t][t][0.8]{0.4}%
		\psfrag{x17}[t][t][0.8]{0.5}%
		\psfrag{x18}[t][t][0.8]{0.6}%
		\psfrag{x19}[t][t][0.8]{0}%
		\psfrag{x20}[t][t][0.8]{0.1}%
		\psfrag{x21}[t][t][0.8]{0.2}%
		\psfrag{x22}[t][t][0.8]{0.3}%
		\psfrag{x23}[t][t][0.8]{0.4}%
		\psfrag{x24}[t][t][0.8]{0.5}%
		\psfrag{x25}[t][t][0.8]{0.6}%
		\psfrag{v01}[r][r][0.8]{0}%
		\psfrag{v02}[r][r][0.8]{0.1}%
		\psfrag{v03}[r][r][0.8]{0.2}%
		\psfrag{v04}[r][r][0.8]{0.3}%
		\psfrag{v05}[r][r][0.8]{0.4}%
		\psfrag{v06}[r][r][0.8]{0.5}%
		\psfrag{v07}[r][r][0.8]{0.6}%
		\psfrag{v08}[r][r][0.8]{0.7}%
		\psfrag{v09}[r][r][0.8]{0.8}%
		\psfrag{v10}[r][r][0.8]{0.9}%
		\psfrag{v11}[r][r][0.8]{1}%
		\psfrag{v12}[r][r][0.8]{-1}%
		\psfrag{v13}[r][r][0.8]{-0.5}%
		\psfrag{v14}[r][r][0.8]{0}%
		\psfrag{v15}[r][r][0.8]{0.5}%
		\psfrag{v16}[r][r][0.8]{1}%
		\psfrag{v17}[r][r][0.8]{-0.15}%
		\psfrag{v18}[r][r][0.8]{-0.1}%
		\psfrag{v19}[r][r][0.8]{-0.05}%
		\psfrag{v20}[r][r][0.8]{0}%
		\psfrag{v21}[r][r][0.8]{0.05}%
		\psfrag{v22}[r][r][0.8]{0.1}%
    \includegraphics[width=10cm]{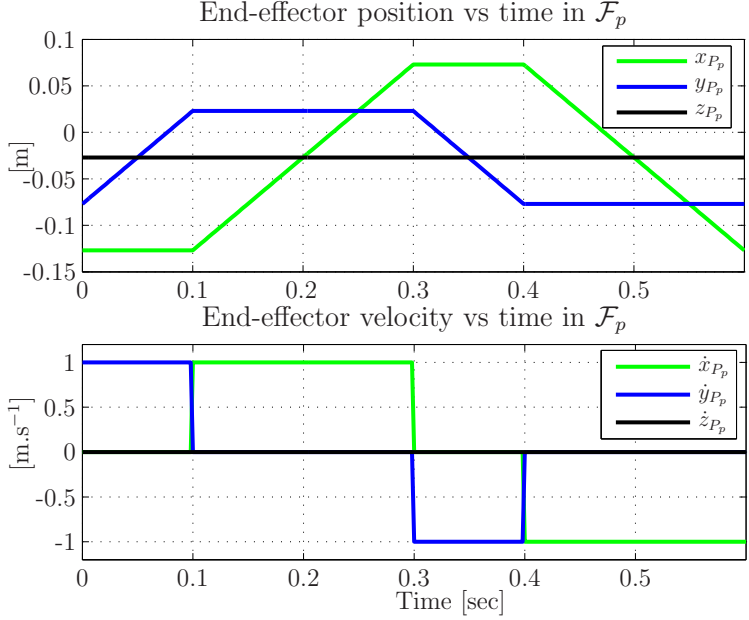}
   \caption{Test trajectory for a rectangular path of size $0.05$m $\times~0.10$m} 
   \label{fig:TestTraj}
\end{figure}

\begin{figure}[h]			
  \centering
  \psfrag{Xb}[c][c][1]{$X_b$}
  \psfrag{Yb}[c][c][1]{$Y_b$}
  \psfrag{Zb}[c][c][1]{$Z_b$}
  \psfrag{Fa}[c][c][1]{\textcolor{myred}{$F_a$}}
  \psfrag{Fr}[c][c][1]{\textcolor{myred}{$F_r$}}
  \psfrag{Ff}[c][c][1]{\textcolor{myred}{$F_f$}}
  \psfrag{Vp}[c][c][1]{$V_p$}
  \includegraphics[width=6cm]{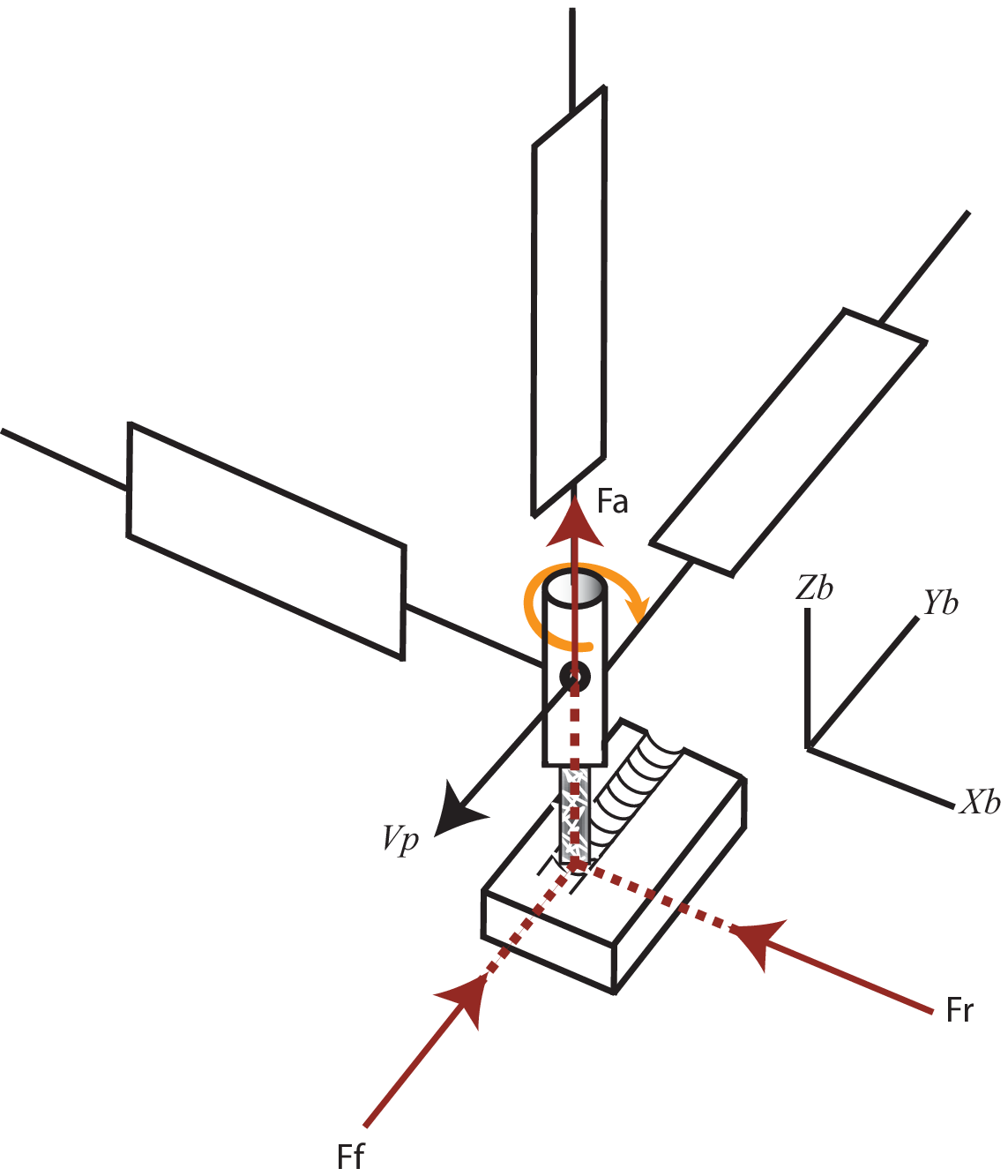}
  \caption{Cutting forces}
  \label{fig:CuttingF}
\end{figure}

In order to apply the methodology proposed for path placement optimization, a rectangular test path is considered. The test path is defined by the length $L$ and the width $W$ of the rectangle, as shown in Fig.~\ref{fig:TestPath}($a$). Path reference frame ${\cal F}_p$ is located at the geometric center of the rectangle. This type of path can be the example of the generation of a rectangular pocket like that of Fig.~\ref{fig:TestPath}($b$). The position of ${\cal F}_p$ in the base frame ${\cal F}_b$ is defined with the Cartesian coordinates of the origin of ${\cal F}_p$, $O_p$($x_{O_p}$, $y_{O_p}$, $z_{O_p}$) and the orientation of ${\cal F}_p$ with respect to ${\cal F}_b$ is given by Euler's angles, as depicted in Fig.~\ref{fig:FbFpFrames}($b$). For the sake of simplicity, only one of the three rotation angles is considered i.e, rotation about $Z_b$-axis while $X_bY_b$ and $X_pY_p$ planes are considered to be always parallel. Accordingly, there are four path placement variables, i.e., $x_{O_p}$, $y_{O_p}$, $z_{O_p}$ and $\phi$, as illustrated in Fig.~\ref{fig:TestPath}.
The magnitude of the end-effector velocity is supposed to be constant along the path. Hence, for given path dimensions, position vector $\mathbf{p_{{\cal F}_p}}=[x_{Pp} \quad y_{Pp} \quad z_{Pp}]^T$ and velocity vector $\mathbf{v_{{\cal F}_p}}=[\dot{x}_{Pp} \quad \dot{y}_{Pp} \quad \dot{z}_{Pp}]^T$ in the path frame can be evaluated as a function of time. Figure \ref{fig:TestTraj} shows the position and velocity profiles in ${\cal F}_p$ for a $0.05\times0.10$ m rectangular path and for a constant end-effector velocity of $1.0$~m.s$^{-1}$.
Position and velocity vectors defined in ${\cal F}_p$ can be expressed in ${\cal F}_b$ by means of the transformation matrix defined in Eq.~(\ref{Eq:bTp}), namely,\\
	\begin{center}
	$\left[	\begin{array}{c}
			x_{P_b}\\
			y_{P_b}\\
			z_{P_b}\\
			1
				\end{array}
	\right]$
	=
	$\left[ \begin{array}{cccc}
  							\cos{\phi}	&	-\sin{\phi}	&  0	&	x_{Op}		\\
  							\sin{\phi}	&	\cos{\phi}	&  0	&	y_{Op}		\\
  								0					&		0					&	 1	&	z_{Op}		\\
  								0					&		0					&	 0	&	1		\\
				\end{array}
	\right] $
	$\left[	\begin{array}{c}
			x_{P_p}\\
			y_{P_p}\\
			z_{P_p}\\
			1
				\end{array}
	\right]$ \\ 
	$\left[	\begin{array}{c}
			\dot{x}_{P_b}\\
			\dot{y}_{P_b}\\
			\dot{z}_{P_b}\\
			1
				\end{array}
	\right]$
	=
	$\left[ \begin{array}{cccc}
  							\cos{\phi}	&	-\sin{\phi}	&  0	&	\quad0		\\
  							\sin{\phi}	&	\cos{\phi}	&  0	&	\quad0		\\
  								0					&		0					&	 1	&	\quad0		\\
  								0					&		0					&	 0	&	\quad1		\\
				\end{array}
	\right] $
	$\left[	\begin{array}{c}
			\dot{x}_{P_p}\\
			\dot{y}_{P_p}\\
			\dot{z}_{P_p}\\
			1
				\end{array}
	\right]$	\\
\end{center}
with $\mathbf{x}=[x_{O_p} \quad y_{O_p}\quad z_{O_p} \quad \phi ]^T$ being the decision variables vector of the optimization problem. For a matter of simplicity and not to deal with tangent and curvature discontinuities, we consider that the path is composed of four independent line segments. Therefore, we do not pay attention to the discontinuities between the segments.\\
In order to analyze the effect of external cutting/machining forces in the generation of a given path, a groove milling operation is considered as shown in Fig. \ref{fig:CuttingF}, \cite{Majou2007}. With constant feed rate or end-effector velocity $\textit{v}_p$ of magnitude $0.66$~m.s$^{-1}$, i.e, $40$~ m.min$^{-1}$, the following components of cutting forces are considered:\\
$F_f$=component in the feed direction = 10 N\\
$F_a$=component along the axis of cutting tool = 25 N\\
$F_r$=component perpendicular to $F_f$ and $F_a$= 215 N\

\subsection{Path Placement Optimization Problem}
The path placement optimization problem for the Orthoglide 3-axis is formulated in order to minimize the total electric energy used by the three prismatic actuators. The kinematic, velocity and dynamic models of the manipulator are used to evaluate the required actuator displacements, velocities and torques. The constraints of the optimization problem are the geometric, kinematic and dynamic ones. It should be noted that within the prescribed workspace, the Orthoglide 3-axis is free of internal collisions and that there is not any limit on the passive joints. Therefore, the geometric constraints for the path placement problem are the upper and lower limits of the prismatic joints variables. The kinematic constraints are the maximum velocities that the prismatic actuators can produce whereas the dynamic constraints are the limits of the torque/force that the actuators can produce. The kinematic and dynamic constraints are obtained from the catalogue, as shown in Table~\ref{Tab:OrthoConstr}. As already mentioned, the decision variables are the Cartesian coordinates of the origin of ${\cal F}_p$ and the orientation angle of ${\cal F}_p$ with respect to ${\cal F}_b$.\\ 
The optimization problem can be formulated as follows, 	
\begin{equation}		
\min_{\mathbf{x}} E_t=\sum^n_{i=1}{E_i}\left( \mathbf{x} \right) \qquad
\textrm{ subject to:} \left\{ 
									\begin{array}{l l}
  								\rho_{min}\leq \rho_{x,y,z}\leq \rho_{max}\\
									\left| v_{x,y,z} \right| \leq v_{max}\\		
									\left| \tau_{x,y,z} \right| \leq \tau_{max}	\qquad \qquad 
									\end{array} \right. 
\label{Eq:MOO_Ortho}
\end{equation}	
where $\mathbf{x}=[x_{Op}, y_{Op}, z_{Op},\phi]^T$. The subscripts $x$, $y$ and $z$ are used for three prismatic actuators or three Cartesian directions. $\rho_{min}$ and $\rho_{max}$ are respectively the minimum and maximum displacements of the prismatic joints as presented in Table~\ref{Tab:OrthoWS}\\
The optimization problem was solved by using the MATLAB \textit{fmincon} function, which is a general constrained optimization solver using the derivative-based search algorithms. 
The optimization process was performed with different starting points and it turned out that MATLAB \textit{fmincon} function always converges to the same solution no matter the starting point. 
Furthermore, to study the variation pattern of energy requirements at different points within the workspace, a workspace discretisation is carried out with respect to the path placement variables and the energy is calculated for each of the discrete point for a given path while verifying the constraints.
\subsection{Results and Discussions of the Path Placement Optimization Problem}
\begin{figure}[h]	
  \centering
	\psfrag{s01}[lt][lt][0.8]{\color[rgb]{0,0,0}\setlength{\tabcolsep}{0pt}\begin{tabular}{l}$X_b$~[mm]\end{tabular}}%
		\psfrag{s02}[rt][rt][0.8]{\color[rgb]{0,0,0}\setlength{\tabcolsep}{0pt}\begin{tabular}{r}$Y_b$~[mm]\end{tabular}}%
		\psfrag{s03}[b][b][0.8]{\color[rgb]{0,0,0}\setlength{\tabcolsep}{0pt}\begin{tabular}{c}$Z_b$~[mm]\end{tabular}}%
		\psfrag{s04}[r][r][0.8]{\color[rgb]{0,0.49804,0}\setlength{\tabcolsep}{0pt}\begin{tabular}{r}$C$\end{tabular}}%
		\psfrag{s05}[r][r][0.8]{\color[rgb]{0,0,1}\setlength{\tabcolsep}{0pt}\begin{tabular}{r}$O_b$\end{tabular}}%
		\psfrag{s06}[r][r][0.8]{\color[rgb]{1,0,0}\setlength{\tabcolsep}{0pt}\begin{tabular}{r}$Q^+$\end{tabular}}%
		\psfrag{s07}[l][l][0.8]{\color[rgb]{1,0,0}\setlength{\tabcolsep}{0pt}\begin{tabular}{l}$Q^-$\end{tabular}}%
		\psfrag{s08}[r][r][0.8]{\color[rgb]{0,0,1}\setlength{\tabcolsep}{0pt}\begin{tabular}{r}$Y_b$\end{tabular}}%
		\psfrag{s09}[r][r][0.8]{\color[rgb]{0,0,1}\setlength{\tabcolsep}{0pt}\begin{tabular}{r} $Z_b$\end{tabular}}%
		\psfrag{s11}[r][r][0.8]{\color[rgb]{0,0,1}\setlength{\tabcolsep}{0pt}\begin{tabular}{r}$X_b$\end{tabular}}%
		\psfrag{s16}[t][t][0.8]{\color[rgb]{0,0,0}\setlength{\tabcolsep}{0pt}\begin{tabular}{c}$20\times40$\end{tabular}}%
		\psfrag{s17}[t][t][0.8]{\color[rgb]{0,0,1}\setlength{\tabcolsep}{0pt}\begin{tabular}{c}$30\times60$\end{tabular}}%
	\psfrag{s22}[b][b][0.8]{\color[rgb]{0.48,0.063,0.894}\setlength{\tabcolsep}{0pt}\begin{tabular}{c}$40\times80$\end{tabular}}%
		\psfrag{s27}[b][b][0.8]{\color[rgb]{0.6,0.2,0}\setlength{\tabcolsep}{0pt}\begin{tabular}{c}$60\times120$\end{tabular}}%
		\psfrag{s28}[b][b][0.8]{\color[rgb]{0,0.49804,0}\setlength{\tabcolsep}{0pt}\begin{tabular}{c}$70\times140$\end{tabular}}%
		\psfrag{s29}[b][b][0.8]{\color[rgb]{1,0,0}\setlength{\tabcolsep}{0pt}\begin{tabular}{c}$80\times160$\end{tabular}}%
		\psfrag{s30}[b][b][0.8]{\color[rgb]{1,0,1}\setlength{\tabcolsep}{0pt}\begin{tabular}{c}$50\times100$\end{tabular}}%
		\psfrag{x01}[t][t][0.8]{0}%
		\psfrag{x02}[t][t][0.8]{0.1}%
		\psfrag{x03}[t][t][0.8]{0.2}%
		\psfrag{x04}[t][t][0.8]{0.3}%
		\psfrag{x05}[t][t][0.8]{0.4}%
		\psfrag{x06}[t][t][0.8]{0.5}%
		\psfrag{x07}[t][t][0.8]{0.6}%
		\psfrag{x08}[t][t][0.8]{0.7}%
		\psfrag{x09}[t][t][0.8]{0.8}%
		\psfrag{x10}[t][t][0.8]{0.9}%
		\psfrag{x11}[t][t][0.8]{1}%
		\psfrag{x12}[t][t][0.8]{-100}%
		\psfrag{x13}[t][t][0.8]{-50}%
		\psfrag{x14}[t][t][0.8]{0}%
		\psfrag{x15}[t][t][0.8]{50}%
		\psfrag{v01}[r][r][0.8]{0}%
		\psfrag{v02}[r][r][0.8]{0.1}%
		\psfrag{v03}[r][r][0.8]{0.2}%
		\psfrag{v04}[r][r][0.8]{0.3}%
		\psfrag{v05}[r][r][0.8]{0.4}%
		\psfrag{v06}[r][r][0.8]{0.5}%
		\psfrag{v07}[r][r][0.8]{0.6}%
		\psfrag{v08}[r][r][0.8]{0.7}%
		\psfrag{v09}[r][r][0.8]{0.8}%
		\psfrag{v10}[r][r][0.8]{0.9}%
		\psfrag{v11}[r][r][0.8]{1}%
		\psfrag{v12}[r][r][0.8]{-100}%
		\psfrag{v13}[r][r][0.8]{-50}%
		\psfrag{v14}[r][r][0.8]{0}%
		\psfrag{v15}[r][r][0.8]{50}%
		\psfrag{z01}[r][r][0.8]{-120}%
		\psfrag{z02}[r][r][0.8]{-100}%
		\psfrag{z03}[r][r][0.8]{-80}%
		\psfrag{z04}[r][r][0.8]{-60}%
		\psfrag{z05}[r][r][0.8]{-40}%
		\psfrag{z06}[r][r][0.8]{-20}%
		\psfrag{z07}[r][r][0.8]{0}%
		\psfrag{z08}[r][r][0.8]{20}%
		\psfrag{z09}[r][r][0.8]{40}%
		\psfrag{z10}[r][r][0.8]{60}%
	 \includegraphics[width=10cm]{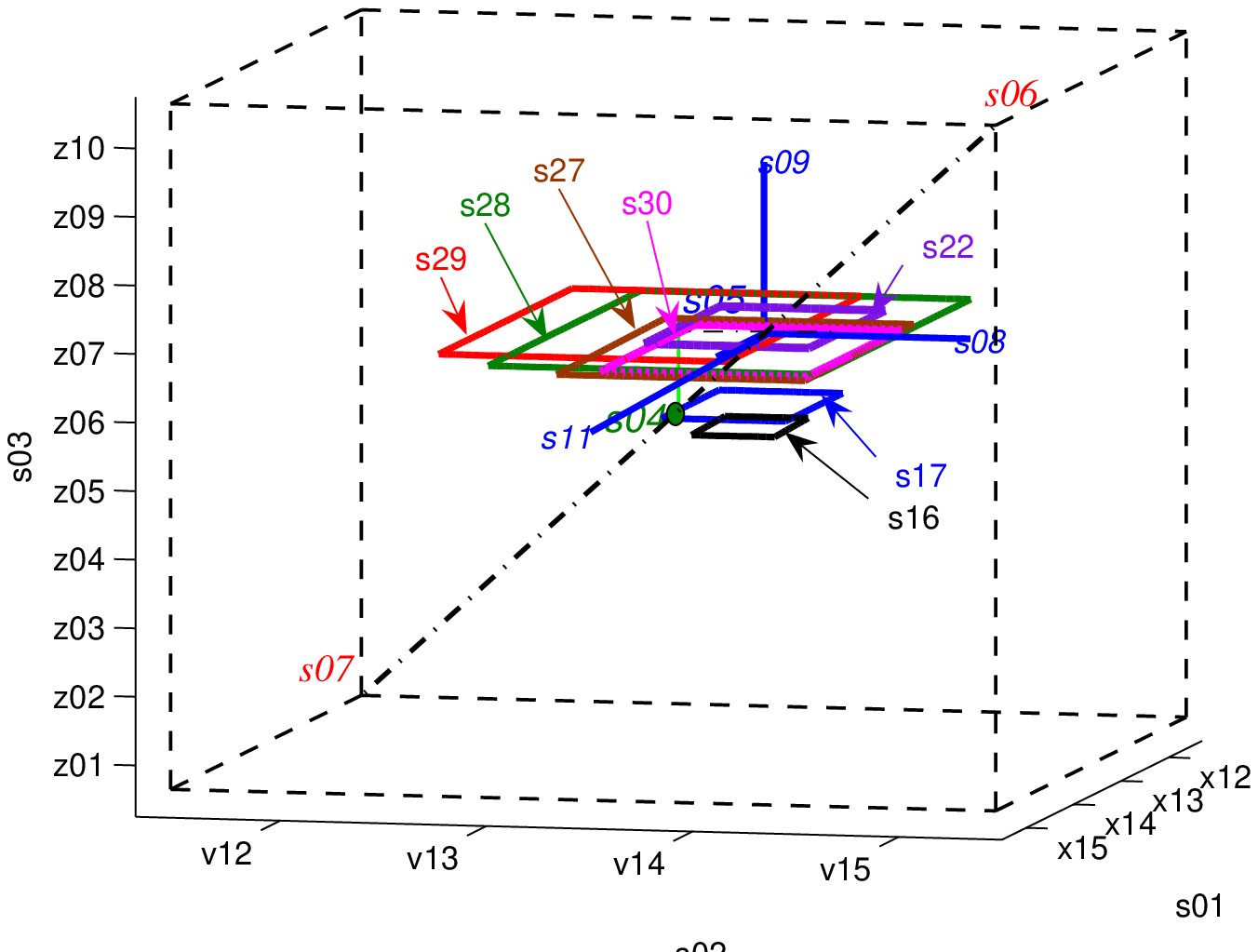}
   \caption{Locations of rectangular path of different sizes ($W$mm~$\times~L$mm) that yield a minimum energy consumption }
   \label{fig:MinETraj}
\end{figure}

\begin{figure}[!htbp]		
  \centering
		\psfrag{s01}[lt][lt][0.8]{\color[rgb]{0,0,0}\setlength{\tabcolsep}{0pt}\begin{tabular}{l}$X_b$~[mm]\end{tabular}}%
		\psfrag{s02}[rt][rt][0.8]{\color[rgb]{0,0,0}\setlength{\tabcolsep}{0pt}\begin{tabular}{r}$Y_b$~[mm]\end{tabular}}%
		\psfrag{s03}[b][b][0.8]{\color[rgb]{0,0,0}\setlength{\tabcolsep}{0pt}\begin{tabular}{c}$Z_b$~[mm]\end{tabular}}%
		\psfrag{s04}[r][r][0.8]{\color[rgb]{0,0.49804,0}\setlength{\tabcolsep}{0pt}\begin{tabular}{r}$C$\end{tabular}}%
		\psfrag{s05}[r][r][0.75]{\color[rgb]{0,0,1}\setlength{\tabcolsep}{0pt}\begin{tabular}{r}$O_b$\end{tabular}}%
		\psfrag{s06}[r][r][0.8]{\color[rgb]{1,0,0}\setlength{\tabcolsep}{0pt}\begin{tabular}{r}$Q^+$\end{tabular}}%
		\psfrag{s07}[l][l][0.8]{\color[rgb]{1,0,0}\setlength{\tabcolsep}{0pt}\begin{tabular}{l}$Q^-$\end{tabular}}%
		\psfrag{s08}[r][r][0.8]{\color[rgb]{0,0,1}\setlength{\tabcolsep}{0pt}\begin{tabular}{r}$X_b$\end{tabular}}%
		\psfrag{s09}[r][r][0.8]{\color[rgb]{0,0,1}\setlength{\tabcolsep}{0pt}\begin{tabular}{r}$Y_b$\end{tabular}}%
		\psfrag{s10}[r][r][0.8]{\color[rgb]{0,0,1}\setlength{\tabcolsep}{0pt}\begin{tabular}{r}$Z_b$\end{tabular}}%
		\psfrag{s20}[b][b][0.7]{\color[rgb]{1,0,0}\setlength{\tabcolsep}{0pt}\begin{tabular}{c}$80\times160$\end{tabular}}%
		\psfrag{s21}[b][b][0.7]{\color[rgb]{0,0.49804,0}\setlength{\tabcolsep}{0pt}\begin{tabular}{c}$70\times140$\end{tabular}}%
		\psfrag{s22}[b][b][0.7]{\color[rgb]{0.6,0.2,0}\setlength{\tabcolsep}{0pt}\begin{tabular}{c}$60\times120$\end{tabular}}%
		\psfrag{s23}[b][b][0.7]{\color[rgb]{1,0,1}\setlength{\tabcolsep}{0pt}\begin{tabular}{c}$50\times100$\end{tabular}}%
			\psfrag{s24}[b][b][0.7]{\color[rgb]{0.48,0.063,0.894}\setlength{\tabcolsep}{0pt}\begin{tabular}{c}$40\times80$\end{tabular}}%
		\psfrag{s25}[b][b][0.7]{\color[rgb]{0,0,1}\setlength{\tabcolsep}{0pt}\begin{tabular}{c}$30\times60$\end{tabular}}%
		\psfrag{s26}[b][b][0.7]{\color[rgb]{0,0,0}\setlength{\tabcolsep}{0pt}\begin{tabular}{c}$20\times40$\end{tabular}}%
		\psfrag{x01}[t][t][0.8]{0}%
		\psfrag{x02}[t][t][0.8]{0.1}%
		\psfrag{x03}[t][t][0.8]{0.2}%
		\psfrag{x04}[t][t][0.8]{0.3}%
		\psfrag{x05}[t][t][0.8]{0.4}%
		\psfrag{x06}[t][t][0.8]{0.5}%
		\psfrag{x07}[t][t][0.8]{0.6}%
		\psfrag{x08}[t][t][0.8]{0.7}%
		\psfrag{x09}[t][t][0.8]{0.8}%
		\psfrag{x10}[t][t][0.8]{0.9}%
		\psfrag{x11}[t][t][0.8]{1}%
		\psfrag{x12}[t][t][0.8]{-100}%
		\psfrag{x13}[t][t][0.8]{-50}%
		\psfrag{x14}[t][t][0.8]{0}%
		\psfrag{x15}[t][t][0.8]{50}%
		\psfrag{v01}[r][r][0.8]{0}%
		\psfrag{v02}[r][r][0.8]{0.1}%
		\psfrag{v03}[r][r][0.8]{0.2}%
		\psfrag{v04}[r][r][0.8]{0.3}%
		\psfrag{v05}[r][r][0.8]{0.4}%
		\psfrag{v06}[r][r][0.8]{0.5}%
		\psfrag{v07}[r][r][0.8]{0.6}%
		\psfrag{v08}[r][r][0.8]{0.7}%
		\psfrag{v09}[r][r][0.8]{0.8}%
		\psfrag{v10}[r][r][0.8]{0.9}%
		\psfrag{v11}[r][r][0.8]{1}%
		\psfrag{v12}[r][r][0.8]{-100}%
		\psfrag{v13}[r][r][0.8]{-50}%
		\psfrag{v14}[r][r][0.8]{0}%
		\psfrag{v15}[r][r][0.8]{50}%
		\psfrag{z01}[r][r][0.8]{-120}%
		\psfrag{z02}[r][r][0.8]{-100}%
		\psfrag{z03}[r][r][0.8]{-80}%
		\psfrag{z04}[r][r][0.8]{-60}%
		\psfrag{z05}[r][r][0.8]{-40}%
		\psfrag{z06}[r][r][0.8]{-20}%
		\psfrag{z07}[r][r][0.8]{0}%
		\psfrag{z08}[r][r][0.8]{20}%
		\psfrag{z09}[r][r][0.8]{40}%
		\psfrag{z10}[r][r][0.8]{60}%

	 \includegraphics[width=10cm]{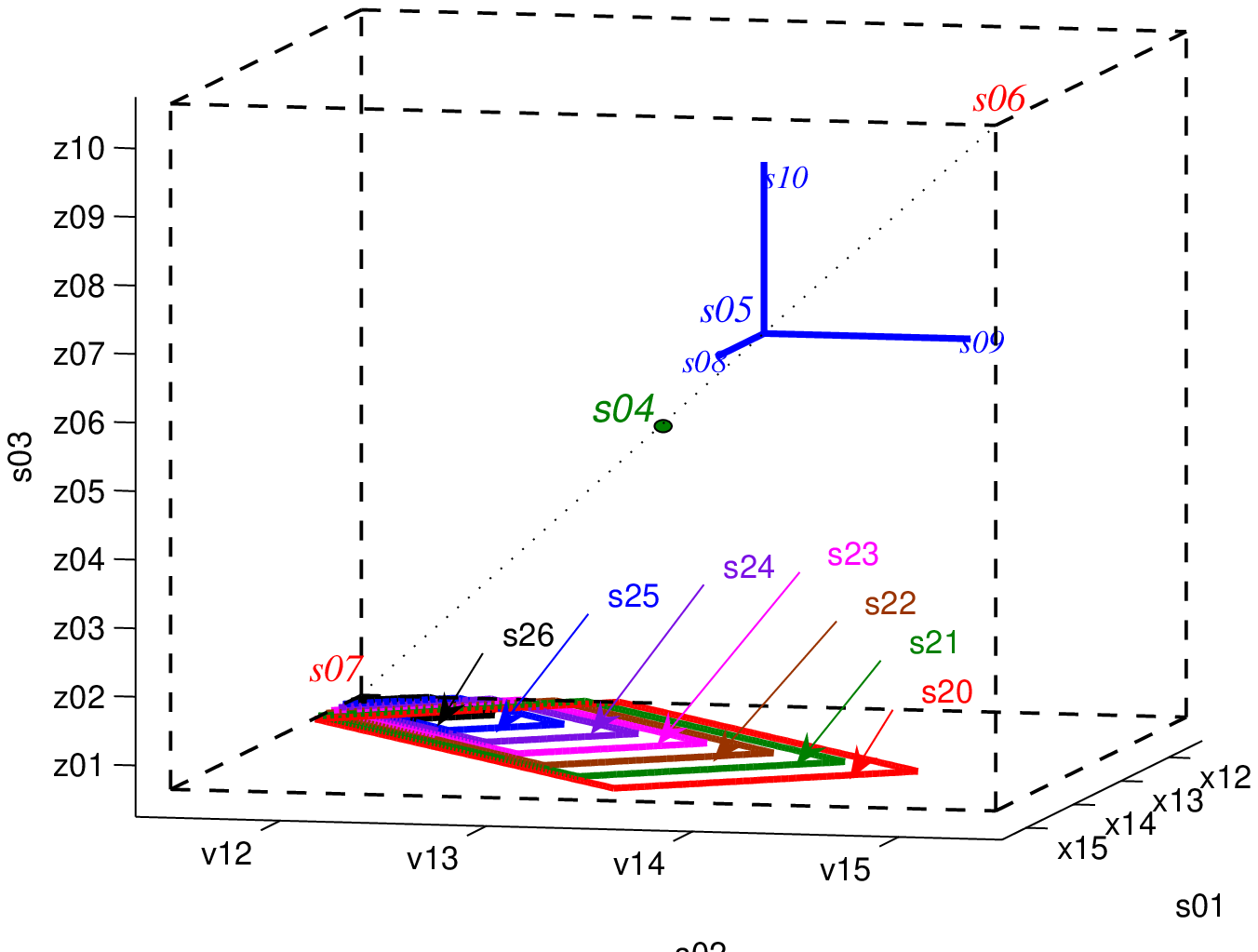}
   \caption{Locations of rectangular path of different sizes ($W$mm~$\times~L$mm) that yield a maximum energy consumption}
   \label{fig:EmaxTraj}
\end{figure}

\begin{figure}[h]		
  \centering
		\psfrag{s05}[b][b][0.8]{\color[rgb]{0,0,0}\setlength{\tabcolsep}{0pt}\begin{tabular}{c}$E$~[J]\end{tabular}}%
		\psfrag{s06}[b][b][0.8]{\color[rgb]{0,0,0}\setlength{\tabcolsep}{0pt}\begin{tabular}{c}Energy ($E$) used for different rectangular path widths\end{tabular}}%
		\psfrag{s10}[t][t][0.8]{\color[rgb]{0,0,0}\setlength{\tabcolsep}{0pt}\begin{tabular}{c}Path widths $W$ [mm]\end{tabular}}%
		\psfrag{s11}[b][b][0.8]{\color[rgb]{0,0,0}\setlength{\tabcolsep}{0pt}\begin{tabular}{c}\%saving\end{tabular}}%
		\psfrag{s12}[b][b][0.8]{\color[rgb]{0,0,0}\setlength{\tabcolsep}{0pt}\begin{tabular}{c}Percentage saving for different rectangular path widths\end{tabular}}%
		\psfrag{s15}[l][l][0.7]{\color[rgb]{0,0,0}$E_{min}$}%
		\psfrag{s16}[l][l][0.7]{\color[rgb]{0,0,0}$E_{max}$}%
		\psfrag{x01}[t][t][0.8]{0}%
		\psfrag{x02}[t][t][0.8]{0.1}%
		\psfrag{x03}[t][t][0.8]{0.2}%
		\psfrag{x04}[t][t][0.8]{0.3}%
		\psfrag{x05}[t][t][0.8]{0.4}%
		\psfrag{x06}[t][t][0.8]{0.5}%
		\psfrag{x07}[t][t][0.8]{0.6}%
		\psfrag{x08}[t][t][0.8]{0.7}%
		\psfrag{x09}[t][t][0.8]{0.8}%
		\psfrag{x10}[t][t][0.8]{0.9}%
		\psfrag{x11}[t][t][0.8]{1}%
		\psfrag{x12}[t][t][0.8]{20}%
		\psfrag{x13}[t][t][0.8]{30}%
		\psfrag{x14}[t][t][0.8]{40}%
		\psfrag{x15}[t][t][0.8]{50}%
		\psfrag{x16}[t][t][0.8]{60}%
		\psfrag{x17}[t][t][0.8]{70}%
		\psfrag{x18}[t][t][0.8]{80}%
		\psfrag{x19}[t][t][0.8]{20}%
		\psfrag{x20}[t][t][0.8]{30}%
		\psfrag{x21}[t][t][0.8]{40}%
		\psfrag{x22}[t][t][0.8]{50}%
		\psfrag{x23}[t][t][0.8]{60}%
		\psfrag{x24}[t][t][0.8]{70}%
		\psfrag{x25}[t][t][0.8]{80}%
		\psfrag{v01}[r][r][0.8]{0}%
		\psfrag{v02}[r][r][0.8]{0.1}%
		\psfrag{v03}[r][r][0.8]{0.2}%
		\psfrag{v04}[r][r][0.8]{0.3}%
		\psfrag{v05}[r][r][0.8]{0.4}%
		\psfrag{v06}[r][r][0.8]{0.5}%
		\psfrag{v07}[r][r][0.8]{0.6}%
		\psfrag{v08}[r][r][0.8]{0.7}%
		\psfrag{v09}[r][r][0.8]{0.8}%
		\psfrag{v10}[r][r][0.8]{0.9}%
		\psfrag{v11}[r][r][0.8]{1}%
		\psfrag{v12}[r][r][0.8]{40}%
		\psfrag{v13}[r][r][0.8]{45}%
		\psfrag{v14}[r][r][0.8]{50}%
		\psfrag{v15}[r][r][0.8]{55}%
		\psfrag{v16}[r][r][0.8]{60}%
		\psfrag{v17}[r][r][0.8]{65}%
		\psfrag{v18}[r][r][0.8]{70}%
		\psfrag{v19}[r][r][0.8]{0}%
		\psfrag{v20}[r][r][0.8]{20}%
		\psfrag{v21}[r][r][0.8]{40}%
		\psfrag{v22}[r][r][0.8]{60}%
		\psfrag{v23}[r][r][0.8]{80}%
		\psfrag{v24}[r][r][0.8]{100}%
		\psfrag{v25}[r][r][0.8]{120}%
		\psfrag{v26}[r][r][0.8]{140}%
	 \includegraphics[width=9cm]{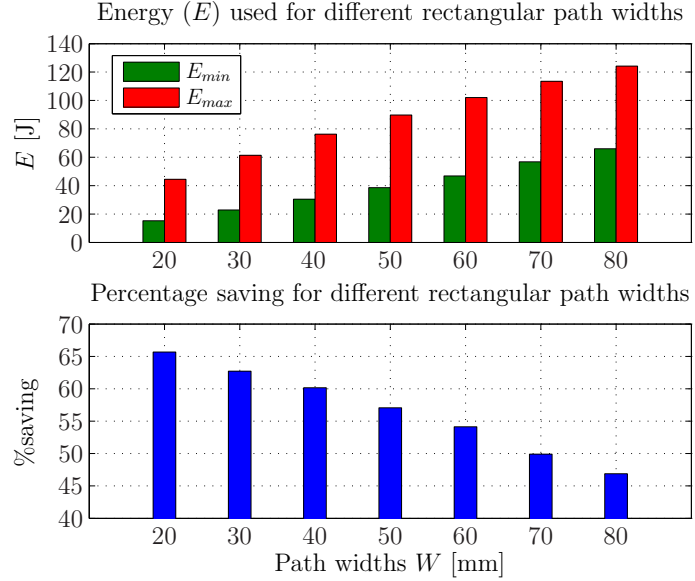}
   \caption{$E_{min}$ and $E_{max}$ and percentage saving as a function of the rectangular path width ($L=2W$)}
   \label{fig:EmaxEminEco}
\end{figure}

The path placement process introduced in this paper is highlighted by means of the rectangular path shown in Fig.~\ref{fig:TestPath}. As a matter of fact, the path placement process is performed for different rectangles with constant aspect ratio of 2,  i.e., $L/W=2$. With the help of the optimization algorithm, the location of the path corresponding to minimum and maximum energy consumption is obtained, i.e., the best and the worst path locations with respect to the electric energy consumed. Figures \ref{fig:MinETraj} and \ref{fig:EmaxTraj} show the location of different rectangular paths with the minimum and maximum energy consumption in the Orthoglide 3-axis cubic workspace. The magnitude of the energy used for both best and worst cases and the corresponding gain of energy is given in Table \ref{Tab:EminEmax} and is illustrated in Fig.~\ref{fig:EmaxEminEco}. In Fig.~\ref{fig:EmaxEminEco}, \% saving is the percent energy saving between the best(minimum) and the worst(maximum) energy consumption.
	\begin{table}[htbp]			
	\centering
		\begin{tabular}{|c|c|c|c|c|}
		\hline
		\multicolumn{2}{|c|}{\centering Rectangular path dimensions  [mm]} &	\multirow{2}{15mm}{\centering $E_{min}$ [J]}	&	\multirow{2}{15mm}{\centering $E_{max}$ [J]}	& \multirow{2}{15mm}{\centering $\%$ gain}\\	 
		\cline{1-2}
		\quad\quad Width ($W$) \quad\quad 	 & Length ($L$)	&		&		&\\
		\hline
				$ 20  $	& $40$	&	$15.26$		&	$44.46$		&	$65.68$\\
		\hline
				$ 30 $	& $60$	&	$22.88$		&	$61.35$		&	$62.71$\\
		\hline
				$ 40 $	& $80$	&	$30.41$		&	$76.31$		&	$60.15$\\
		\hline
				$ 50 $	& $100$	&	$38.55$		&	$89.80$		&	$57.07$\\
		\hline
				$ 60 $	& $120$	&	$46.83$		&	$102.11$		&	$54.13$\\
		\hline
				$ 70 $	& $140$	&	$56.82$		&	$113.46$		&	$49.92$\\
		\hline
				$ 80 $	& $160$	&	$65.94$		&	$121.17$		&	$46.89$\\
		\hline
		\end{tabular}
	\caption{Minimum and maximum energy used for a given rectangular path}
	\label{Tab:EminEmax}
\end{table}
It can be seen from Figs.~\ref{fig:MinETraj} and \ref{fig:EmaxTraj} that the energy consumption is a minimum when the path is located in the vicinity of the isotropic configuration with velocity and force transmission factors equal to one, i.e. point $O_b$, with $\phi=0\,^{\circ}$ and is a maximum when the path is located in the vicinity of point $Q^-$ with $\phi=45\,^{\circ}$. From Fig.~\ref{fig:EmaxEminEco}, it can be noticed that the smaller the path, the higher the energy saving. This higher gain for the smaller path is due to the higher range of displacement of the path within the manipulator workspace.\\
\begin{figure}[!htbp]		
  \centering
		\psfrag{s05}[lt][lt][0.8]{\color[rgb]{0,0,0}\setlength{\tabcolsep}{0pt}\begin{tabular}{l}$x_{Op}$~[mm]\end{tabular}}%
		\psfrag{s06}[rt][rt][0.8]{\color[rgb]{0,0,0}\setlength{\tabcolsep}{0pt}\begin{tabular}{r}$z_{Op}$~[mm]\end{tabular}}%
		\psfrag{s07}[b][b][0.8]{\color[rgb]{0,0,0}\setlength{\tabcolsep}{0pt}\begin{tabular}{c}$E$~[J]\end{tabular}}%
		\psfrag{s08}[b][b][0.8]{\color[rgb]{0,0,0}\setlength{\tabcolsep}{0pt}\begin{tabular}{c}$E$ vs $x_{Op}z_{Op}$~ ($\phi=0\,^{\circ}$, $W=30$~mm, $L=60$~mm)\end{tabular}}%
		\psfrag{s24}[t][t][0.8]{\color[rgb]{1,0,1}\setlength{\tabcolsep}{0pt}\begin{tabular}{l}$y_{Op}=0$~mm\end{tabular}}%
		\psfrag{s25}[b][b][0.8]{\color[rgb]{0,0.49804,0}\setlength{\tabcolsep}{0pt}\begin{tabular}{l}$y_{Op}=-102$~mm\end{tabular}}%
		\psfrag{s26}[b][b][0.8]{\color[rgb]{0,0,1}\setlength{\tabcolsep}{0pt}\begin{tabular}{l}$y_{Op}=48$~mm\end{tabular}}%
		\psfrag{x01}[t][t][0.8]{0}%
		\psfrag{x02}[t][t][0.8]{0.1}%
		\psfrag{x03}[t][t][0.8]{0.2}%
		\psfrag{x04}[t][t][0.8]{0.3}%
		\psfrag{x05}[t][t][0.8]{0.4}%
		\psfrag{x06}[t][t][0.8]{0.5}%
		\psfrag{x07}[t][t][0.8]{0.6}%
		\psfrag{x08}[t][t][0.8]{0.7}%
		\psfrag{x09}[t][t][0.8]{0.8}%
		\psfrag{x10}[t][t][0.8]{0.9}%
		\psfrag{x11}[t][t][0.8]{1}%
		\psfrag{x12}[t][t][0.8]{0}%
		\psfrag{x13}[t][t][0.8]{0.1}%
		\psfrag{x14}[t][t][0.8]{0.2}%
		\psfrag{x15}[t][t][0.8]{0.3}%
		\psfrag{x16}[t][t][0.8]{0.4}%
		\psfrag{x17}[t][t][0.8]{0.5}%
		\psfrag{x18}[t][t][0.8]{0.6}%
		\psfrag{x19}[t][t][0.8]{0.7}%
		\psfrag{x20}[t][t][0.8]{0.8}%
		\psfrag{x21}[t][t][0.8]{0.9}%
		\psfrag{x22}[t][t][0.8]{1}%
		\psfrag{x23}[t][t][0.8]{-100}%
		\psfrag{x24}[t][t][0.8]{-50}%
		\psfrag{x25}[t][t][0.8]{0}%
		\psfrag{x26}[t][t][0.8]{50}%
		\psfrag{v01}[r][r][0.8]{0}%
		\psfrag{v02}[r][r][0.8]{0.1}%
		\psfrag{v03}[r][r][0.8]{0.2}%
		\psfrag{v04}[r][r][0.8]{0.3}%
		\psfrag{v05}[r][r][0.8]{0.4}%
		\psfrag{v06}[r][r][0.8]{0.5}%
		\psfrag{v07}[r][r][0.8]{0.6}%
		\psfrag{v08}[r][r][0.8]{0.7}%
		\psfrag{v09}[r][r][0.8]{0.8}%
		\psfrag{v10}[r][r][0.8]{0.9}%
		\psfrag{v11}[r][r][0.8]{1}%
		\psfrag{v12}[l][l][0.8]{24}%
		\psfrag{v13}[l][l][0.8]{26}%
		\psfrag{v14}[l][l][0.8]{28}%
		\psfrag{v15}[l][l][0.8]{30}%
		\psfrag{v16}[l][l][0.8]{32}%
		\psfrag{v17}[l][l][0.8]{34}%
		\psfrag{v18}[l][l][0.8]{36}%
		\psfrag{v19}[l][l][0.8]{38}%
		\psfrag{v20}[l][l][0.8]{40}%
		\psfrag{v21}[r][r][0.8]{-200}%
		\psfrag{v22}[r][r][0.8]{-100}%
		\psfrag{v23}[r][r][0.8]{0}%
		\psfrag{v24}[r][r][0.8]{100}%
		\psfrag{z01}[r][r][0.8]{20}%
		\psfrag{z02}[r][r][0.8]{25}%
		\psfrag{z03}[r][r][0.8]{30}%
		\psfrag{z04}[r][r][0.8]{35}%
		\psfrag{z05}[r][r][0.8]{40}%
		\psfrag{z06}[r][r][0.8]{45}%
	 \includegraphics[width=9cm]{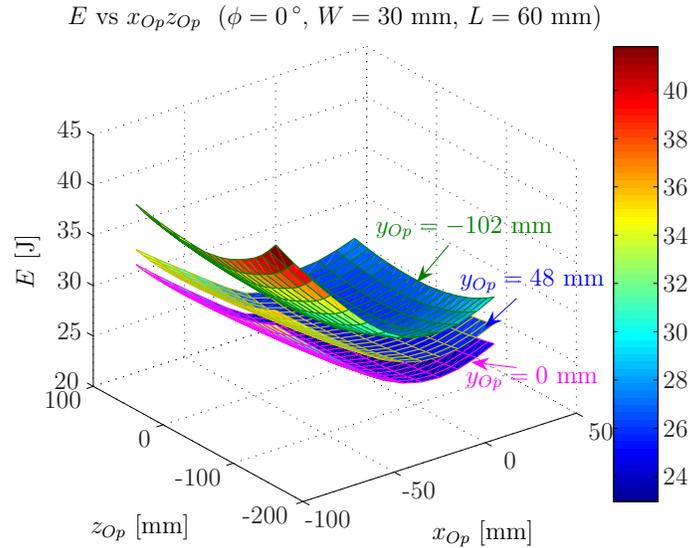}
   \caption{Energy as a function of $x_{Op}$ and $z_{Op}$ for a 30mm$\times$60mm  rectangular path}
   \label{fig:ExzYconst}
\end{figure}
\begin{figure}[htbp]			
  \centering
  	\psfrag{s01}[][][0.8]{\color[rgb]{0,0,0}\setlength{\tabcolsep}{0pt}\begin{tabular}{c}23\end{tabular}}%
		\psfrag{s02}[][][0.8]{\color[rgb]{0,0,0}\setlength{\tabcolsep}{0pt}\begin{tabular}{c}23\end{tabular}}%
		\psfrag{s03}[][][0.8]{\color[rgb]{0,0,0}\setlength{\tabcolsep}{0pt}\begin{tabular}{c}24\end{tabular}}%
		\psfrag{s04}[][][0.8]{\color[rgb]{0,0,0}\setlength{\tabcolsep}{0pt}\begin{tabular}{c}24\end{tabular}}%
		\psfrag{s05}[][][0.8]{\color[rgb]{0,0,0}\setlength{\tabcolsep}{0pt}\begin{tabular}{c}24\end{tabular}}%
		\psfrag{s06}[][][0.8]{\color[rgb]{0,0,0}\setlength{\tabcolsep}{0pt}\begin{tabular}{c}24\end{tabular}}%
		\psfrag{s07}[][][0.8]{\color[rgb]{0,0,0}\setlength{\tabcolsep}{0pt}\begin{tabular}{c}24\end{tabular}}%
		\psfrag{s08}[][][0.8]{\color[rgb]{0,0,0}\setlength{\tabcolsep}{0pt}\begin{tabular}{c}25\end{tabular}}%
		\psfrag{s09}[][][0.8]{\color[rgb]{0,0,0}\setlength{\tabcolsep}{0pt}\begin{tabular}{c}25\end{tabular}}%
		\psfrag{s10}[][][0.8]{\color[rgb]{0,0,0}\setlength{\tabcolsep}{0pt}\begin{tabular}{c}25\end{tabular}}%
		\psfrag{s11}[][][0.8]{\color[rgb]{0,0,0}\setlength{\tabcolsep}{0pt}\begin{tabular}{c}25\end{tabular}}%
		\psfrag{s12}[][][0.8]{\color[rgb]{0,0,0}\setlength{\tabcolsep}{0pt}\begin{tabular}{c}25\end{tabular}}%
		\psfrag{s13}[][][0.8]{\color[rgb]{0,0,0}\setlength{\tabcolsep}{0pt}\begin{tabular}{c}26\end{tabular}}%
		\psfrag{s14}[][][0.8]{\color[rgb]{0,0,0}\setlength{\tabcolsep}{0pt}\begin{tabular}{c}26\end{tabular}}%
		\psfrag{s15}[][][0.8]{\color[rgb]{0,0,0}\setlength{\tabcolsep}{0pt}\begin{tabular}{c}26\end{tabular}}%
		\psfrag{s16}[][][0.8]{\color[rgb]{0,0,0}\setlength{\tabcolsep}{0pt}\begin{tabular}{c}26\end{tabular}}%
		\psfrag{s17}[][][0.8]{\color[rgb]{0,0,0}\setlength{\tabcolsep}{0pt}\begin{tabular}{c}26\end{tabular}}%
		\psfrag{s18}[][][0.8]{\color[rgb]{0,0,0}\setlength{\tabcolsep}{0pt}\begin{tabular}{c}28\end{tabular}}%
		\psfrag{s19}[][][0.8]{\color[rgb]{0,0,0}\setlength{\tabcolsep}{0pt}\begin{tabular}{c}28\end{tabular}}%
		\psfrag{s20}[][][0.8]{\color[rgb]{0,0,0}\setlength{\tabcolsep}{0pt}\begin{tabular}{c}28\end{tabular}}%
		\psfrag{s21}[][][0.8]{\color[rgb]{0,0,0}\setlength{\tabcolsep}{0pt}\begin{tabular}{c}30\end{tabular}}%
		\psfrag{s22}[][][0.8]{\color[rgb]{0,0,0}\setlength{\tabcolsep}{0pt}\begin{tabular}{c}30\end{tabular}}%
		\psfrag{s23}[][][0.8]{\color[rgb]{0,0,0}\setlength{\tabcolsep}{0pt}\begin{tabular}{c}30\end{tabular}}%
		\psfrag{s24}[][][0.8]{\color[rgb]{0,0,0}\setlength{\tabcolsep}{0pt}\begin{tabular}{c}32\end{tabular}}%
		\psfrag{s25}[][][0.8]{\color[rgb]{0,0,0}\setlength{\tabcolsep}{0pt}\begin{tabular}{c}32\end{tabular}}%
		\psfrag{s26}[][][0.8]{\color[rgb]{0,0,0}\setlength{\tabcolsep}{0pt}\begin{tabular}{c}32\end{tabular}}%
		\psfrag{s27}[][][0.8]{\color[rgb]{0,0,0}\setlength{\tabcolsep}{0pt}\begin{tabular}{c}34\end{tabular}}%
		\psfrag{s28}[b][b][0.8]{\color[rgb]{0,0,0}\setlength{\tabcolsep}{0pt}\begin{tabular}{c}Energy [J] variation with $x_{Op}$ and $y_{Op}$~ ($z_{Op}=0$,~$\phi=0\,^{\circ}$)\end{tabular}}%
		\psfrag{s29}[t][t][0.8]{\color[rgb]{0,0,0}\setlength{\tabcolsep}{0pt}\begin{tabular}{c}$x_{Op}$~[mm]\end{tabular}}%
		\psfrag{s30}[b][b][0.8]{\color[rgb]{0,0,0}\setlength{\tabcolsep}{0pt}\begin{tabular}{c}$y_{Op}$~[mm]\end{tabular}}%
		\psfrag{x01}[t][t][0.8]{0}%
		\psfrag{x02}[t][t][0.8]{0.1}%
		\psfrag{x03}[t][t][0.8]{0.2}%
		\psfrag{x04}[t][t][0.8]{0.3}%
		\psfrag{x05}[t][t][0.8]{0.4}%
		\psfrag{x06}[t][t][0.8]{0.5}%
		\psfrag{x07}[t][t][0.8]{0.6}%
		\psfrag{x08}[t][t][0.8]{0.7}%
		\psfrag{x09}[t][t][0.8]{0.8}%
		\psfrag{x10}[t][t][0.8]{0.9}%
		\psfrag{x11}[t][t][0.8]{1}%
		\psfrag{x12}[t][t][0.8]{-80}%
		\psfrag{x13}[t][t][0.8]{-60}%
		\psfrag{x14}[t][t][0.8]{-40}%
		\psfrag{x15}[t][t][0.8]{-20}%
		\psfrag{x16}[t][t][0.8]{0}%
		\psfrag{x17}[t][t][0.8]{20}%
		\psfrag{v01}[r][r][0.8]{0}%
		\psfrag{v02}[r][r][0.8]{0.1}%
		\psfrag{v03}[r][r][0.8]{0.2}%
		\psfrag{v04}[r][r][0.8]{0.3}%
		\psfrag{v05}[r][r][0.8]{0.4}%
		\psfrag{v06}[r][r][0.8]{0.5}%
		\psfrag{v07}[r][r][0.8]{0.6}%
		\psfrag{v08}[r][r][0.8]{0.7}%
		\psfrag{v09}[r][r][0.8]{0.8}%
		\psfrag{v10}[r][r][0.8]{0.9}%
		\psfrag{v11}[r][r][0.8]{1}%
		\psfrag{v12}[r][r][0.8]{-100}%
		\psfrag{v13}[r][r][0.8]{-80}%
		\psfrag{v14}[r][r][0.8]{-60}%
		\psfrag{v15}[r][r][0.8]{-40}%
		\psfrag{v16}[r][r][0.8]{-20}%
		\psfrag{v17}[r][r][0.8]{0}%
		\psfrag{v18}[r][r][0.8]{20}%
		\psfrag{v19}[r][r][0.8]{40}%
	 \includegraphics[width=9cm]{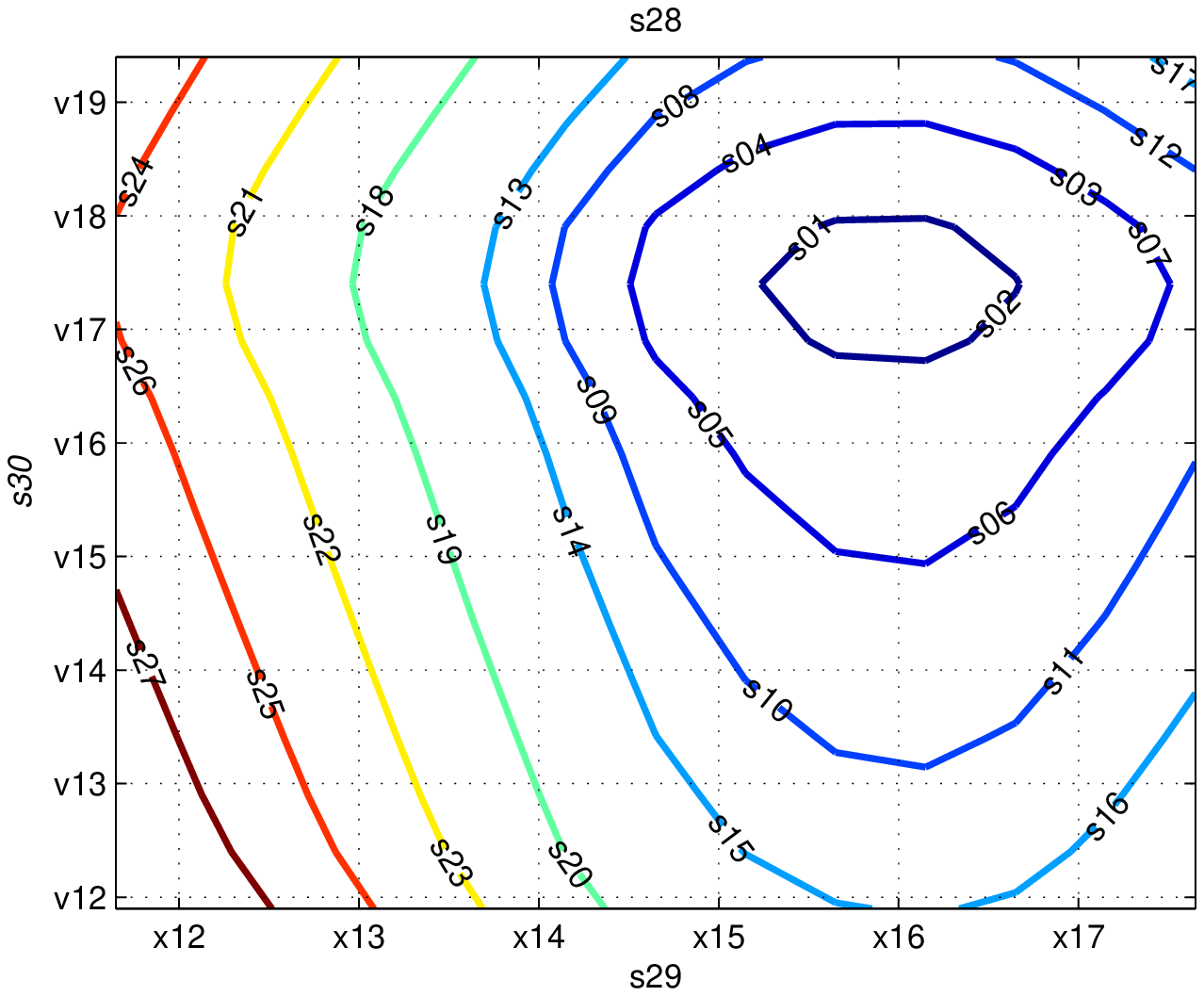}
   \caption{Energy vs $x_{Op}$ and $y_{Op}$ for $30$mm~$\times~60$mm rectangular path}
   \label{fig:IsoContour}
\end{figure}
In order to view the energy variation trends in the workspace, a test path of size $30$mm~$\times$~$60$mm is taken and the energy required for the generation of this test path is evaluated for several positions and orientations. Figure~\ref{fig:ExzYconst} shows the variations in the energy used with respect to $x_{Op}$ and $z_{Op}$ for a constant orientation $\phi$ of $0\,^{\circ}$ and for three different values of  $y_{Op}$. 
Figure~\ref{fig:IsoContour} illustrates the isocontours of the energy required by the motors with respect to $x_{Op}$ and $y_{Op}$ for  given values of $z_{Op}$ and $\phi$, namely, $z_{Op}=0$  and $\phi=0\,^{\circ}$.
From Fig.~\ref{fig:ExzYconst}, it is apparent that the energy required is not sensitive to variations in $z_{Op}$. The reason being that the path lies in the $X_bY_b$-plane and the actuator displacements along $Z_b$-axis is not significant. Figures~\ref{fig:ExzYconst} and \ref{fig:IsoContour} also show that the energy required by the motors is a minimum when the path is located in the neighbourhood of the isotropic configuration and is a maximum when the latter is located in the neighbourhood of singularities.\\ 
\begin{figure}[!htbp]			
  \centering
\psfrag{s05}[lt][lt][0.8]{\color[rgb]{0,0,0}\setlength{\tabcolsep}{0pt}\begin{tabular}{c}$y_{Op}$ [mm]\end{tabular}}%
\psfrag{s06}[r][c][0.8]{\color[rgb]{0,0,0}\setlength{\tabcolsep}{0pt}\begin{tabular}{r}$x_{Op}$ [mm]\end{tabular}}%
\psfrag{s07}[b][b][0.8]{\color[rgb]{0,0,0}\setlength{\tabcolsep}{0pt}\begin{tabular}{c}$E$ [J]\end{tabular}}%
\psfrag{s08}[b][b][0.8]{\color[rgb]{0,0,0}\setlength{\tabcolsep}{0pt}\begin{tabular}{c}$E$ vs $x_{Op}y_{Op}$ ($z_{Op}=0$)\end{tabular}}%
\psfrag{s26}[][][0.8]{\setlength{\fboxsep}{2pt}\fcolorbox[rgb]{0,0,0}{1,0.69412,0.39216}{\color[rgb]{0,0.74902,0.74902}\setlength{\tabcolsep}{0pt}\begin{tabular}{c}$\phi=60\,^{\circ}$\end{tabular}}}%
\psfrag{s27}[][][0.8]{\setlength{\fboxsep}{2pt}\fcolorbox[rgb]{0,0,0}{1,0.69412,0.39216}{\color[rgb]{1,0,1}\setlength{\tabcolsep}{0pt}\begin{tabular}{c}$\phi=45\,^{\circ}$\end{tabular}}}%
\psfrag{s28}[b][b][0.8]{\setlength{\fboxsep}{2pt}\fcolorbox[rgb]{0,0,0}{1,0.69412,0.39216}{\color[rgb]{0,0.49804,0}\setlength{\tabcolsep}{0pt}\begin{tabular}{c}$\phi=30\,^{\circ}$\end{tabular}}}%
\psfrag{s29}[][][0.8]{\setlength{\fboxsep}{2pt}\fcolorbox[rgb]{0,0,0}{1,0.69412,0.39216}{\color[rgb]{0,0,1}\setlength{\tabcolsep}{0pt}\begin{tabular}{c}$\phi=0\,^{\circ}$\end{tabular}}}%
\psfrag{s38}[][][0.8]{\setlength{\fboxsep}{2pt}\fcolorbox[rgb]{0,0,0}{1,0.69412,0.39216}{\color[rgb]{0,0.49804,0}\setlength{\tabcolsep}{0pt}\begin{tabular}{c}$\phi=90\,^{\circ}$\end{tabular}}}%
		\psfrag{x01}[t][t][0.8]{0}%
		\psfrag{x02}[t][t][0.8]{0.1}%
		\psfrag{x03}[t][t][0.8]{0.2}%
		\psfrag{x04}[t][t][0.8]{0.3}%
		\psfrag{x05}[t][t][0.8]{0.4}%
		\psfrag{x06}[t][t][0.8]{0.5}%
		\psfrag{x07}[t][t][0.8]{0.6}%
		\psfrag{x08}[t][t][0.8]{0.7}%
		\psfrag{x09}[t][t][0.8]{0.8}%
		\psfrag{x10}[t][t][0.8]{0.9}%
		\psfrag{x11}[t][t][0.8]{1}%
		\psfrag{x12}[t][t][0.8]{0}%
		\psfrag{x13}[t][t][0.8]{0.1}%
		\psfrag{x14}[t][t][0.8]{0.2}%
		\psfrag{x15}[t][t][0.8]{0.3}%
		\psfrag{x16}[t][t][0.8]{0.4}%
		\psfrag{x17}[t][t][0.8]{0.5}%
		\psfrag{x18}[t][t][0.8]{0.6}%
		\psfrag{x19}[t][t][0.8]{0.7}%
		\psfrag{x20}[t][t][0.8]{0.8}%
		\psfrag{x21}[t][t][0.8]{0.9}%
		\psfrag{x22}[t][t][0.8]{1}%
		
		\psfrag{x23}[c][c][0.8]{-100}%
		\psfrag{x24}[c][c][0.8]{-50}%
		\psfrag{x25}[c][c][0.8]{0}%
		\psfrag{x26}[c][c][0.8]{50}%
		\psfrag{v01}[r][r][0.8]{0}%
		\psfrag{v02}[r][r][0.8]{0.1}%
		\psfrag{v03}[r][r][0.8]{0.2}%
		\psfrag{v04}[r][r][0.8]{0.3}%
		\psfrag{v05}[r][r][0.8]{0.4}%
		\psfrag{v06}[r][r][0.8]{0.5}%
		\psfrag{v07}[r][r][0.8]{0.6}%
		\psfrag{v08}[r][r][0.8]{0.7}%
		\psfrag{v09}[r][r][0.8]{0.8}%
		\psfrag{v10}[r][r][0.8]{0.9}%
		\psfrag{v11}[r][r][0.8]{1}%
		\psfrag{v12}[l][l][0.8]{25}%
		\psfrag{v13}[l][l][0.8]{30}%
		\psfrag{v14}[l][l][0.8]{35}%
		\psfrag{v15}[l][l][0.8]{40}%
		\psfrag{v16}[l][l][0.8]{45}%
		\psfrag{v17}[l][l][0.8]{50}%
		\psfrag{v18}[r][r][0.8]{}%
		\psfrag{v19}[c][c][0.8]{-50}%
		\psfrag{v20}[c][c][0.8]{0}%
		\psfrag{v21}[c][c][0.8]{50}%
		\psfrag{v22}[c][c][0.8]{100}%
		\psfrag{z01}[r][r][0.8]{20}%
		\psfrag{z02}[r][r][0.8]{25}%
		\psfrag{z03}[r][r][0.8]{30}%
		\psfrag{z04}[r][r][0.8]{35}%
		\psfrag{z05}[r][r][0.8]{40}%
		\psfrag{z06}[r][r][0.8]{45}%
	 \includegraphics[width=9cm]{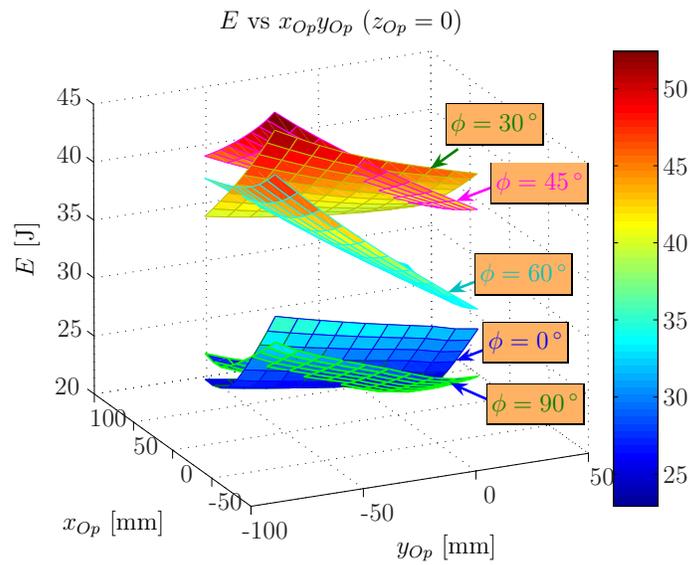}
   \caption{Energy as a function of $x_{Op}$ and $y_{Op}$ for different orientations ($z_{Op}=0$)}
   \label{fig:ExyVsPhi}
\end{figure}
Figures \ref{fig:ExyVsPhi} and \ref{fig:ExPhi} show the variations in the energy used with the path orientation in different areas of the cubic workspace. It can be seen that the energy used is usually a maximum when $\phi=45\,^{\circ}$. However, the energy consumption is a maximum for a path orientation different than $45\,^{\circ}$, for some path locations. For example, the energy required at the upper right corner of the workspace is higher for $\phi=30\,^{\circ}$ than $\phi=45\,^{\circ}$.
\begin{figure}[h]			
  \centering
		\psfrag{s05}[lt][lt][0.8]{\color[rgb]{0,0,0}\setlength{\tabcolsep}{0pt}\begin{tabular}{l}$x_{Op}$ [mm]\end{tabular}}%
		\psfrag{s06}[rt][rt][0.8]{\color[rgb]{0,0,0}\setlength{\tabcolsep}{0pt}\begin{tabular}{r}$\phi$ [deg]\end{tabular}}%
		\psfrag{s07}[b][b][0.8]{\color[rgb]{0,0,0}\setlength{\tabcolsep}{0pt}\begin{tabular}{c}$E$ [J]\end{tabular}}%
		\psfrag{s08}[b][b][0.8]{\color[rgb]{0,0,0}\setlength{\tabcolsep}{0pt}\begin{tabular}{c}$E$ vs $x_{Op}$ and $\phi$ ($y_{Op}=z_{Op}=0$)\end{tabular}}%
		\psfrag{x01}[t][t][0.8]{0}%
		\psfrag{x02}[t][t][0.8]{0.1}%
		\psfrag{x03}[t][t][0.8]{0.2}%
		\psfrag{x04}[t][t][0.8]{0.3}%
		\psfrag{x05}[t][t][0.8]{0.4}%
		\psfrag{x06}[t][t][0.8]{0.5}%
		\psfrag{x07}[t][t][0.8]{0.6}%
		\psfrag{x08}[t][t][0.8]{0.7}%
		\psfrag{x09}[t][t][0.8]{0.8}%
		\psfrag{x10}[t][t][0.8]{0.9}%
		\psfrag{x11}[t][t][0.8]{1}%
		\psfrag{x12}[t][t][0.8]{0}%
		\psfrag{x13}[t][t][0.8]{0.1}%
		\psfrag{x14}[t][t][0.8]{0.2}%
		\psfrag{x15}[t][t][0.8]{0.3}%
		\psfrag{x16}[t][t][0.8]{0.4}%
		\psfrag{x17}[t][t][0.8]{0.5}%
		\psfrag{x18}[t][t][0.8]{0.6}%
		\psfrag{x19}[t][t][0.8]{0.7}%
		\psfrag{x20}[t][t][0.8]{0.8}%
		\psfrag{x21}[t][t][0.8]{0.9}%
		\psfrag{x22}[t][t][0.8]{1}%
		\psfrag{x23}[t][t][0.8]{-100}%
		\psfrag{x24}[t][t][0.8]{-80}%
		\psfrag{x25}[t][t][0.8]{-60}%
		\psfrag{x26}[t][t][0.8]{-40}%
		\psfrag{x27}[t][t][0.8]{-20}%
		\psfrag{x28}[t][t][0.8]{0}%
		\psfrag{x29}[t][t][0.8]{20}%
		\psfrag{x30}[t][t][0.8]{40}%
		\psfrag{v01}[r][r][0.8]{0}%
		\psfrag{v02}[r][r][0.8]{0.1}%
		\psfrag{v03}[r][r][0.8]{0.2}%
		\psfrag{v04}[r][r][0.8]{0.3}%
		\psfrag{v05}[r][r][0.8]{0.4}%
		\psfrag{v06}[r][r][0.8]{0.5}%
		\psfrag{v07}[r][r][0.8]{0.6}%
		\psfrag{v08}[r][r][0.8]{0.7}%
		\psfrag{v09}[r][r][0.8]{0.8}%
		\psfrag{v10}[r][r][0.8]{0.9}%
		\psfrag{v11}[r][r][0.8]{1}%
		\psfrag{v12}[l][l][0.8]{24}%
		\psfrag{v13}[l][l][0.8]{26}%
		\psfrag{v14}[l][l][0.8]{28}%
		\psfrag{v15}[l][l][0.8]{30}%
		\psfrag{v16}[l][l][0.8]{32}%
		\psfrag{v17}[l][l][0.8]{34}%
		\psfrag{v18}[l][l][0.8]{36}%
		\psfrag{v19}[l][l][0.8]{38}%
		\psfrag{v20}[l][l][0.8]{40}%
		\psfrag{v21}[l][l][0.8]{42}%
		\psfrag{v22}[l][l][0.8]{44}%
		\psfrag{v23}[r][r][0.8]{0}%
		\psfrag{v24}[r][r][0.8]{15}%
		\psfrag{v25}[r][r][0.8]{30}%
		\psfrag{v26}[r][r][0.8]{45}%
		\psfrag{v27}[r][r][0.8]{60}%
		\psfrag{v28}[r][r][0.8]{75}%
		\psfrag{v29}[r][r][0.8]{90}%
		\psfrag{v30}[r][r][0.8]{100}%
		\psfrag{z01}[r][r][0.8]{20}%
		\psfrag{z02}[r][r][0.8]{25}%
		\psfrag{z03}[r][r][0.8]{30}%
		\psfrag{z04}[r][r][0.8]{35}%
		\psfrag{z05}[r][r][0.8]{40}%
		\psfrag{z06}[r][r][0.8]{45}%
	 \includegraphics[width=9cm]{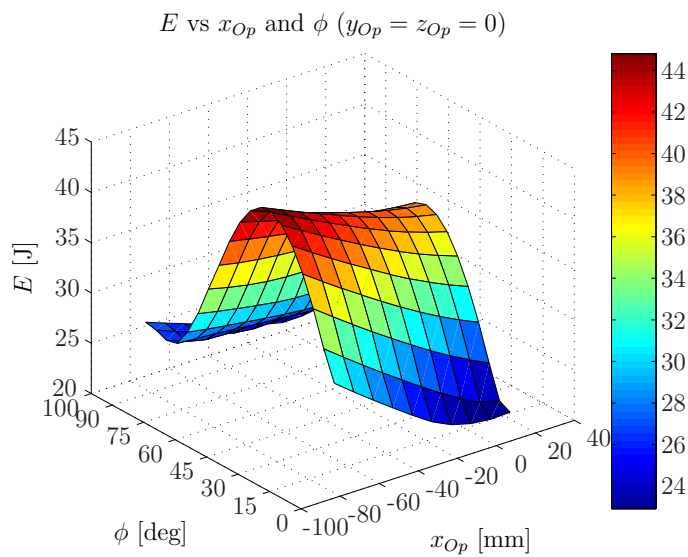}
   \caption{Energy as a function of $x_{Op}$ and $\phi$ for $y_{Op}=z_{Op}=0$}
   \label{fig:ExPhi}
\end{figure}
\begin{figure}[!htbp]			
  \centering
		\psfrag{s05}[b][b][0.8]{\color[rgb]{0,0,0}\setlength{\tabcolsep}{0pt}\begin{tabular}{c}[m]\end{tabular}}%
		\psfrag{s06}[b][b][0.8]{\color[rgb]{0,0,0}\setlength{\tabcolsep}{0pt}\begin{tabular}{c}Max actuator displacements\end{tabular}}%
		\psfrag{s07}[t][t][1.0]{\color[rgb]{0,0,0.5}\setlength{\tabcolsep}{0pt}\begin{tabular}{c}$(a)$\end{tabular}}%
		\psfrag{s10}[b][b][0.8]{\color[rgb]{0,0,0}\setlength{\tabcolsep}{0pt}\begin{tabular}{c}[m/s]\end{tabular}}%
		\psfrag{s11}[b][b][0.8]{\color[rgb]{0,0,0}\setlength{\tabcolsep}{0pt}\begin{tabular}{c}Max actuator velocities\end{tabular}}%
		\psfrag{s12}[t][t][1.0]{\color[rgb]{0,0,0.5}\setlength{\tabcolsep}{0pt}\begin{tabular}{c}$(b)$\end{tabular}}%
		\psfrag{s15}[b][b][0.8]{\color[rgb]{0,0,0}\setlength{\tabcolsep}{0pt}\begin{tabular}{c}[Nm]\end{tabular}}%
		\psfrag{s16}[c][c][0.8]{\color[rgb]{0,0,0}\setlength{\tabcolsep}{0pt}\begin{tabular}{c}Max actuator torques\end{tabular}}%
		\psfrag{s17}[t][t][1.0]{\color[rgb]{0,0,0.5}\setlength{\tabcolsep}{0pt}\begin{tabular}{c}$(c)$\end{tabular}}%
		\psfrag{s20}[b][b][0.8]{\color[rgb]{0,0,0}\setlength{\tabcolsep}{0pt}\begin{tabular}{c}[J]\end{tabular}}%
		\psfrag{s21}[b][b][0.8]{\color[rgb]{0,0,0}\setlength{\tabcolsep}{0pt}\begin{tabular}{c}Energy used by each actuator\end{tabular}}%
		\psfrag{s22}[t][t][1.0]{\color[rgb]{0,0,0.5}\setlength{\tabcolsep}{0pt}\begin{tabular}{c} $(d)$\end{tabular}}%
		\psfrag{s25}[l][l][0.8]{\color[rgb]{0,0,0}$E_{max}$}%
		\psfrag{s26}[l][l][0.8]{\color[rgb]{0,0,0}$E_{min}$}%
		\psfrag{s27}[l][l][0.8]{\color[rgb]{0,0,0}$E_{max}$}%
		\psfrag{x01}[t][t][0.8]{0}%
		\psfrag{x02}[t][t][0.8]{0.1}%
		\psfrag{x03}[t][t][0.8]{0.2}%
		\psfrag{x04}[t][t][0.8]{0.3}%
		\psfrag{x05}[t][t][0.8]{0.4}%
		\psfrag{x06}[t][t][0.8]{0.5}%
		\psfrag{x07}[t][t][0.8]{0.6}%
		\psfrag{x08}[t][t][0.8]{0.7}%
		\psfrag{x09}[t][t][0.8]{0.8}%
		\psfrag{x10}[t][t][0.8]{0.9}%
		\psfrag{x11}[t][t][0.8]{1}%
		\psfrag{x12}[t][t][0.8]{$E_x$}%
		\psfrag{x13}[t][t][0.8]{$E_y$}%
		\psfrag{x14}[t][t][0.8]{$E_z$}%
		\psfrag{x15}[t][t][0.8]{$\tau_x$}%
		\psfrag{x16}[t][t][0.8]{$\tau_y$}%
		\psfrag{x17}[t][t][0.8]{$\tau_z$}%
		\psfrag{x18}[t][t][0.8]{$V_x$}%
		\psfrag{x19}[t][t][0.8]{$V_y$}%
		\psfrag{x20}[t][t][0.8]{$V_z$}%
		\psfrag{x21}[t][t][0.8]{$X$}%
		\psfrag{x22}[t][t][0.8]{$Y$}%
		\psfrag{x23}[t][t][0.8]{$Z$}%
		\psfrag{v01}[r][r][0.8]{0}%
		\psfrag{v02}[r][r][0.8]{0.1}%
		\psfrag{v03}[r][r][0.8]{0.2}%
		\psfrag{v04}[r][r][0.8]{0.3}%
		\psfrag{v05}[r][r][0.8]{0.4}%
		\psfrag{v06}[r][r][0.8]{0.5}%
		\psfrag{v07}[r][r][0.8]{0.6}%
		\psfrag{v08}[r][r][0.8]{0.7}%
		\psfrag{v09}[r][r][0.8]{0.8}%
		\psfrag{v10}[r][r][0.8]{0.9}%
		\psfrag{v11}[r][r][0.8]{1}%
		\psfrag{v12}[r][r][0.8]{0}%
		\psfrag{v13}[r][r][0.8]{10}%
		\psfrag{v14}[r][r][0.8]{20}%
		\psfrag{v15}[r][r][0.8]{30}%
		\psfrag{v16}[r][r][0.8]{40}%
		\psfrag{v17}[r][r][0.8]{50}%
		\psfrag{v18}[r][r][0.8]{0}%
		\psfrag{v19}[r][r][0.8]{0.25}%
		\psfrag{v20}[r][r][0.8]{0.5}%
		\psfrag{v21}[r][r][0.8]{0.75}%
		\psfrag{v22}[r][r][0.8]{1}%
		\psfrag{v23}[r][r][0.8]{1.25}%
		\psfrag{v24}[r][r][0.8]{1.5}%
		\psfrag{v25}[r][r][0.8]{0}%
		\psfrag{v26}[r][r][0.8]{0.2}%
		\psfrag{v27}[r][r][0.8]{0.4}%
		\psfrag{v28}[r][r][0.8]{0.6}%
		\psfrag{v29}[r][r][0.8]{0.8}%
		\psfrag{v30}[r][r][0.8]{0}%
		\psfrag{v31}[r][r][0.8]{0.02}%
		\psfrag{v32}[r][r][0.8]{0.04}%
		\psfrag{v33}[r][r][0.8]{0.06}%
		\psfrag{v34}[r][r][0.8]{0.08}%
	 \includegraphics[width=12cm]{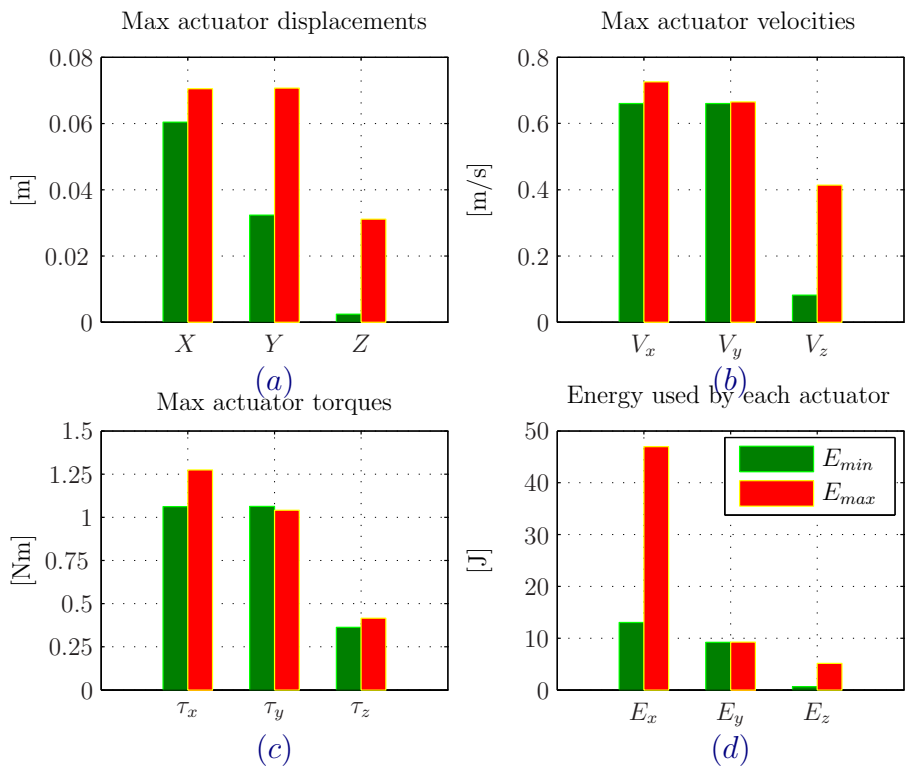}
   \caption{Comparison of $30$mm~$\times$~$60$mm trajectory parameters for $E_{min}$ and $E_{max}$ locations}
   \label{fig:EminMaxTrajPram}
\end{figure}
Figure \ref{fig:EminMaxTrajPram} shows the comparison of trajectory parameters of $30$mm~$\times$~$60$mm test path for minimum and maximum energy locations. It makes sense that when $E=E_{max}$ the range of actuator displacements is larger than when $E=E_{min}$. Similarly actuators experience higher values of maximum velocities and torques when $E=E_{max}$ as shown in Fig.~\ref{fig:EminMaxTrajPram}(b-c), which results in higher energy consumption for each actuator, as shown in Fig.~\ref{fig:EminMaxTrajPram}(d). These results mean that the actuators may reach their performance limits due to an inappropriate location of the path in the workspace.
\section{Acknowledgment}
This work has been partially funded by the European projects NEXT, acronyms for "Next Generation of Productions Systems", Project no° IP 011815. 
\section{CONCLUSIONS}
A methodology for path placement optimization was proposed in this paper. The electric energy required by the actuators to follow a predefined path is considered as the optimization criterion. Electric energy requirement is calculated with the help of the required actuator torques and velocities along with motors electric parameters. To verify the feasibility of the solutions, actuators performance limits such as their joint limits, maximum velocities and torque were used as the constraints of the optimization problem. The kinematic, velocity and dynamic models were used to come up with  the objective function and constraints.

The proposed methodology was applied to the Orthoglide 3-axis, a three-degree-of-freedom translational parallel manipulator with a quasi-cubic workspace. Rectangular test paths of different sizes were considered as illustrative examples. These paths are similar to those used to realize pocketing operations. 

The use of the electric energy instead of mechanical energy as an optimization criterion is pertinent. Although actuator electric energy consumption depends on the mechanical energy requirements, the electric energy evaluation is more comprehensive than its mechanical counterpart. General approach to calculate the mechanical energy with the help of manipulator velocity and dynamic models, i.e., by using actuator torques and velocities, may lead to an under estimation of the energy requirements in the case where actuators are experiencing torques with zero velocities. Besides, usual mechanical energy calculations do not consider the resistive energy loss in the motor windings as well as the energy loss  due to the variations in the actuator velocities. Those variations affect the current requirements and hence induce electromotive forces in the actuators. Accordingly, the electric energy formulation takes into account all these energy losses.

The energy required to perform a given task depends on the position and the orientation of the task within the workspace of the manipulator. Accordingly, some electric energy can be saved by properly selecting the position and the orientation of the task. Indeed, a misplaced task can cause excessive energy consumption and can force the actuators to go over their performance limits.

For the Orthoglide 3-axis, the optimum path location is found to be in the neighbourhood of the isotropic configuration but there is no general rule to predict the exact optimal position and orientation of a task particularly for a complicated three dimensional task or for an irregular workspace. However, a detailed analysis of the energy variation within the workspace for a given task can lead to the optimal position/orientation of that particular task. Numerical optimization algorithms are useful for such a comprehensive analysis in which all the problem constraints and performance measures can be considered simultaneously.

In the future work, the path placement optimization problem will be dealt as a multiobjective one. For example, along with energy requirements, the manipulator dexterity and stiffness can be considered as optimization objectives.

\end{document}